\documentclass{article} 
\PassOptionsToPackage{table}{xcolor}
\usepackage{iclr2027_conference,times}

\usepackage{amsmath,amsfonts,bm}

\def\eqref#1{equation~\ref{#1}}
\def\Eqref#1{Equation~\ref{#1}}

\def\1{\bm{1}}

\DeclareMathAlphabet{\mathsfit}{\encodingdefault}{\sfdefault}{m}{sl}
\SetMathAlphabet{\mathsfit}{bold}{\encodingdefault}{\sfdefault}{bx}{n}

\usepackage{amsmath,amssymb}
\usepackage{hyperref}
\hypersetup{
  hidelinks,
  pageanchor=false
}
\usepackage{url}
\usepackage{graphicx}
\usepackage{tabularx}
\usepackage{array}
\usepackage{booktabs}
\usepackage[table]{xcolor}
\usepackage{multirow} 
\usepackage{fontawesome5}

\usepackage{enumitem}

\title{World-Calibrated Proposal-to-Action Flow for Vision-Language-Action Models}

\author{
Jie He$^{1,*}$, Wei Li$^{1,*}$, Junwen Tong$^{2,*}$, Rui Shao$^{1,\dagger}$,
Wei-Shi Zheng$^{3}$, Liqiang Nie$^{1}$ \\
\textsuperscript{1}Harbin Institute of Technology, Shenzhen \\
\textsuperscript{2}State Key Laboratory of Mobile Network and Mobile Multimedia Technology, ZTE, China \\
\textsuperscript{3}Sun Yat-sen University \\
$^*$Equal Contributions, $^\dagger$Corresponding Author\\
\texttt{shaorui@hit.edu.cn} \vspace{2mm}\\
\faGlobe~\textbf{Homepage}: \texttt{\href{https://JiuTian-VL.github.io/ProAct-page}
{\textcolor{magenta}{https://JiuTian-VL.github.io/ProAct-page}}} \\
\faGithub~\textbf{Github}: \texttt{\href{https://github.com/JiuTian-VL/ProAct}
{\textcolor{magenta}{https://github.com/JiuTian-VL/ProAct}}}
}

\newcommand{\method}{\textsc{ProAct}}
\newcommand{\sg}{\operatorname{sg}}
\newcommand{\Attn}{\operatorname{Attn}}

\newcommand{\BlockDiag}{\operatorname{BlockDiag}}

\iclrfinalcopy 

\begin{document}

\maketitle

\begin{abstract}
Flow-based Vision-Language-Action (VLA) policies generate action chunks by transporting samples from a task-agnostic isotropic Gaussian source. As this source is conditioned on neither recent execution nor predicted future evolution, \textit{\textbf{(i)}} it discards the local continuity established by recently executed motion. \textit{\textbf{(ii)}} Even when predictive world representations are introduced, they often only condition the transport dynamics rather than determine where generation starts, how far it may deviate, or along which action directions it may expand.
Building on this observation, we introduce \textbf{\textsc{ProAct}}, a world-calibrated proposal-to-action framework that makes the generative source itself predictable. 
\textit{\textbf{(i)}} To preserve motion continuity, a lightweight Proposal Expert converts recent actions into a scene-aware hypothesis via one motion-anchored endpoint flow-matching step, initializing generation near the demonstrated action manifold.
\textit{\textbf{(ii)}} To jointly capture intended scene evolution and proposal--future compatibility, a prospective World Expert treats the hypothesis as a soft motion prior while predicting the task-consistent latent future.
\textit{\textbf{(iii)}} From this compatibility, the model calibrates a proposal-centered anisotropic source, where a bounded per-step extent controls the allowed deviation and a trace-normalized low-rank geometry under a condition-number budget allocates refinement over coupled translation, rotation, and gripper directions.
Compared with $\pi_{0.5}$, \textbf{\textsc{ProAct}} improves performance across simulation and real-world tasks while reducing denoising steps by 50\%, inference latency by up to 25.8\%, and increasing throughput by up to 34.8\%.
\end{abstract}

\vspace{-6pt}
\section{Introduction}
\label{sec:introduction}
\vspace{-3pt}

\begin{figure}[t]
\vspace{-4mm}
\centering
\includegraphics[width=1.0\textwidth]{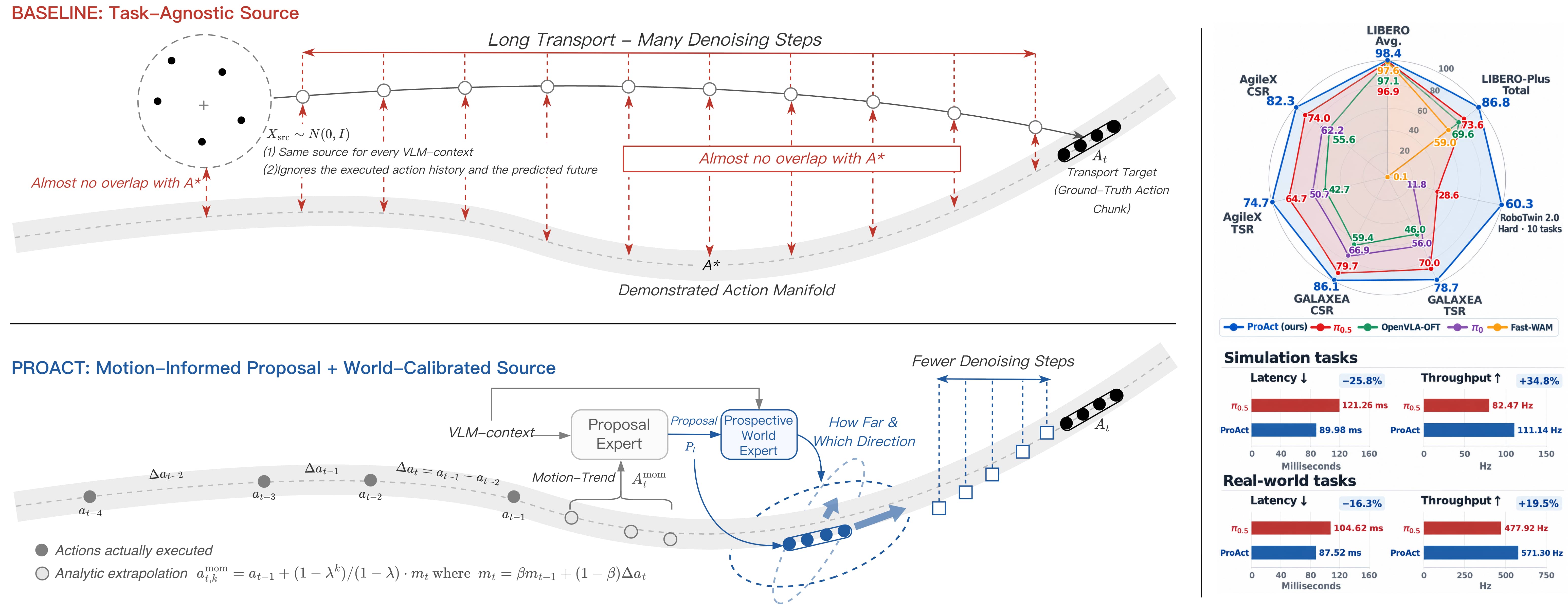}
\vspace{-6mm}
\caption{
\textbf{Motivation of \textsc{ProAct}}. \textsc{ProAct} uses recent motion to form a scene-aware proposal and prospective world reasoning to calibrate its source distribution. Starting from this informed, proposal-centered source shortens action transport while improving performance and efficiency.
}
\vspace{-3mm}
\label{fig:intro}
\end{figure}

Vision-language-action (VLA) models~\cite{pi0, bjorck2025gr00t, li2026causal, luo2026being, shao2025large} increasingly use diffusion~\cite{chi2025diffusion} or flow matching~\cite{lipman2022flow} to generate continuous robot actions, building on broader progress in fine-grained multimodal understanding and visual grounding~\cite{peng2026finegrained, shao2024detecting, shao2023detecting, shao2019multi}. However, these generative decoders typically initialize each action chunk from a generic source distribution~\cite{pi05, zheng2026x}, forcing the policy to repeatedly transport uninformed samples toward executable trajectories, as illustrated in the upper panel of Fig.~\ref{fig:intro}. 
Such a generic source ignores readily available context that could inform where action generation should begin.
By making the source distribution itself contextually predictable, action generation can begin from a more informed region of the action space, yielding
a policy that is both more accurate and substantially cheaper to sample during closed-loop policy execution.

A generic isotropic source leaves two useful sources of information unexploited. 
The first is \textit{\textbf{recently executed motion}}. Although task-level behavior can be highly multimodal, physical motion often exhibits strong local regularity over short horizons, and recently executed, reliable actions are already available during closed-loop control~\cite{black2026real, wang2026real}. 
A naive strategy is to extrapolate this motion and reuse it for the next action chunk~\cite{jia2026action, zhan2026seam}. Yet such continuation becomes unreliable around contact, reorientation, gripper actuation, or task-phase transitions, where appropriate motion may deviate sharply from recent behavior~\cite{yang2026chunkflow}. Local continuity therefore indicates what is \emph{plausible} from the recent past, but not necessarily what remains \emph{appropriate} for the current scene and task. Recent motion should thus be treated as a proposal to be evaluated against the intended future, rather than as the action itself.

The second is \textit{\textbf{predicted future evolution}}, which provides prospective evidence for calibrating such a proposal.
Prospective world modeling~\cite{zhao2025cot, zhang2025up, tian2025predictive, zhang2026dreamvla, li2026drivevla, zhu2026delta, li2026unimpa} has increasingly been incorporated into VLA policies, but typically as auxiliary supervision or additional conditioning.
In these formulations, prospective information helps predict what should happen next, but is not explicitly used to calibrate an existing action hypothesis according to whether it remains appropriate for the anticipated task evolution.
The missing element is therefore an explicit interface between a candidate motion, prospective world reasoning, and generative refinement.

We build this missing interface around a \emph{propose--calibrate--refine} principle and instantiate it as \textbf{\textsc{ProAct}}, a World-Calibrated Proposal-to-Action Flow framework. \textsc{ProAct} first constructs a scene-aware motion hypothesis, then uses prospective world reasoning to estimate the correction required to realize the intended future, and finally converts this calibration into the source distribution for flow-based action generation, as illustrated in the lower panel of Fig.~\ref{fig:intro}.

Specifically, 
\textit{\textbf{(i)}} a lightweight \textit{\textbf{Proposal Expert}} first converts recently executed actions into a scene-aware motion hypothesis. It extrapolates short-horizon motion and uses it to initialize the motion-anchored flow module, which grounds the proposal in the current VLM context through a single endpoint flow step. Rather than producing an executable action, this stage places generation near a plausible region of the demonstrated action manifold and provides a \textit{\textbf{soft motion prior}} for subsequent reasoning. 
\textit{\textbf{(ii)}} A prospective \textit{\textbf{World Expert}} calibrates this proposal against the task-consistent future. The motion hypothesis is provided as contextual evidence for prospective reasoning, while the prediction target remains the demonstrated future encoded in the latent space of a frozen V-JEPA~2 encoder~\cite{assran2025vjepa2}. 
Calibration readouts are further supervised by the residual between the demonstrated action and the proposal, forcing the shared world representation to encode both the task-consistent latent future and the proposal--future compatibility. 
\textit{\textbf{(iii)}} The resulting calibration is finally expressed as a \textit{\textbf{proposal-centered anisotropic source}} for action generation. A bounded per-step \textit{\textbf{extent}} limits how far generation may deviate from the proposal, while trace-normalized low-rank \textit{\textbf{geometry}} determines along which coupled action directions that correction should occur. 
An Action Expert then transports samples from this calibrated source into executable action chunks using the base VLA's original flow parameterization. Thus, \textbf{\textsc{ProAct}} changes where and how action generation begins while preserving the flexibility to depart from the proposal when needed.

Our primary contributions are summarized as follows:
\begin{itemize}[leftmargin=1.2em]

\item We formulate generative action prediction from a propose--calibrate--refine perspective, where recent motion provides a soft local hypothesis calibrated by task-consistent prospective reasoning, providing an interface between local motion continuity and future-aware action generation.

\item We introduce an asymmetric proposal-aware World Expert that uses motion to assist future prediction while anchoring the desired future independently of proposal correctness, and jointly predicts proposal-specific refinement extent and geometry at every horizon step.

\item We replace the isotropic Gaussian of flow-based VLAs with a world-calibrated, proposal-centered source while leaving their transport parameterization unchanged.

\item We validate \textsc{ProAct} across extensive simulation benchmarks and real-world robotic manipulation tasks, where it consistently outperforms strong baselines while reducing denoising steps by up to 50\%, demonstrating both superior effectiveness and inference efficiency.
\end{itemize}

\begin{figure}[t]
\vspace{-4mm}
\centering
\includegraphics[width=1.0\textwidth]{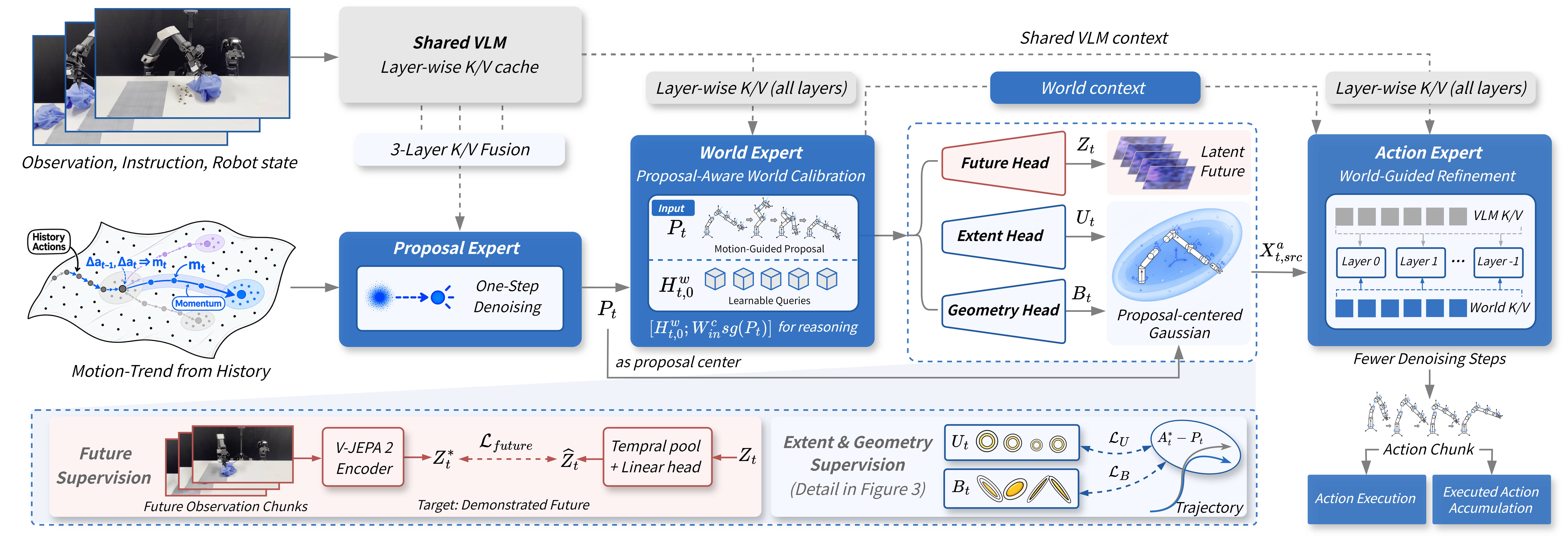}
\vspace{-6mm}
\caption{
\textbf{Overview of \method{}.}
A Proposal Expert forms a scene-aware motion hypothesis, which the World Expert calibrates against the task-consistent future to predict refinement extent and geometry.
The resulting source is then refined into executable actions by the Action Expert.
}
\label{fig:framework}
\end{figure}

\vspace{-6pt}
\section{\method{}}
\label{sec:method}
\vspace{-3pt}

As shown in Fig.~\ref{fig:framework}, \method{} follows a
\emph{propose--calibrate--refine} pipeline. A \textit{\textbf{Proposal Expert}} grounds
recent motion into a scene-aware hypothesis, a prospective \textit{\textbf{World Expert}}
models latent future and proposal--future compatibility, and an
\textit{\textbf{Action Expert}} transports the calibrated source into executable
actions. This calibration determines where refinement starts, how far it may
travel, and along which coupled directions it should proceed at each horizon step.

\vspace{-2pt}
\subsection{Problem Formulation}
\label{sec:problem_formulation}
\vspace{-2pt}

Given demonstrations $\mathcal{D}=\{(\tau_i,\ell_i)\}_{i=1}^{N}$ with
$\tau_i=\{(O_t,q_t,a_t)\}_{t=1}^{T_i}$, we predict an action chunk
$A_t=(a_{t,1},\ldots,a_{t,H})\in\mathbb{R}^{H\times d_a}$ from
$\mathcal{C}_t=(O_{\leq t},\ell,q_{\leq t})$ and recent action history
$\mathcal{H}_t^a=\{a_{t-M},\ldots,a_{t-1}\}$ containing only executed actions.
We use a shared horizon $M=H$. A VLM encodes $\mathcal{C}_t$ and exposes layer-wise
keys and values
$\mathcal{K}_t^{\mathrm{VLM}}
=\{(K_{t,l}^{v},V_{t,l}^{v})\}_{l=1}^{L_v}$,
which provide common perception--language context to all experts throughout generation.

Suppressing time indices and treating $\mathcal{K}_t^{\mathrm{VLM}}$ as
implicit context, for notational clarity we let $C=\mathcal{C}_t$,
$H^a=\mathcal{H}_t^a$, and $X=X_{\mathrm{src}}$ throughout this formulation.
Generation then factorizes as
\begin{equation}
\small
p(A\mid C,H^a)
=
\iiint
\underbrace{p_{\theta}(A\mid X,H^w)}_{\text{\scriptsize Action}}\!
\underbrace{p_{\eta}(X\mid P,H^w)}_{\text{\scriptsize Source}}\!
\underbrace{p_{\omega}(H^w\mid P)}_{\text{\scriptsize World}}\!
\underbrace{p_{\phi}(P\mid H^a)}_{\text{\scriptsize Proposal}}\,
dH^w\,dX\,dP .
\label{eq:proact_factorization}
\end{equation}
Read right-to-left, the factors propose motion, model its desired future,
calibrate the source, and generate actions; $p_\omega$ may condition on $P$ when beneficial, whereas $p_\eta$ must explicitly preserve $P$ as the source anchor and calibrate deviations around it.

\vspace{-2pt}
\subsection{Motion-Informed Action Proposal}
\label{sec:proposal}
\vspace{-2pt}

Robot motion is locally continuous despite task-level multimodality. Let
$\Delta a_t=a_{t-1}-a_{t-2}$ and maintain the smoothed motion
state $m_t=\beta m_{t-1}+(1-\beta)\Delta a_t$. We extrapolate
\begin{equation}
\small
a_{t,k}^{\mathrm{mom}}
=
a_{t-1}
+
\frac{1-\lambda^k}{1-\lambda}m_t,
\quad
k=1,\ldots,H ,
\label{eq:motion_proposal}
\end{equation}
where $\beta$ suppresses fluctuations and $\lambda$ attenuates long-horizon
extrapolation; gripper dimensions retain their last executed value. The chunk
$A_t^{\mathrm{mom}}$ captures local motion continuity but is not scene-aware.
It initializes smooth motion while leaving contact and target changes to visual
grounding.

A shallow Proposal Expert grounds this analytic prior in VLM context. It uses
$L_p$ layers of the Action Expert and a single flow time $t_p=1$, treating
$A_t^{\mathrm{mom}}$ as a source rather than noise. During training, we perturb it as
$X_t^p=A_t^{\mathrm{mom}}+\sigma_p\epsilon_p$,
$\epsilon_p\sim\mathcal{N}(0,I)$, and initialize
$H_{t,0}^p=W_{\mathrm{in}}^pX_t^p$. Each layer jointly attends to VLM context
and the proposal state:
\begin{equation}
\small
\widehat H_{t,j}^p
=
\Attn\!\left(
Q_j^p(H_{t,j-1}^p),
[\widetilde K_{t,j}^v;K_j^p],
[\widetilde V_{t,j}^v;V_j^p]
\right),
\label{eq:proposal_joint_attention}
\end{equation}
where $K_j^p,V_j^p$ project $H_{t,j-1}^p$. Because the proposer is shallower
than the VLM ($L_p=6$ versus $L_v=18$), its layer $j$ receives a compressed
cache from $\mathcal{S}_j=\{3j-2,3j-1,3j\}$. It averages jointly RMS-normalized
caches using parameter-free reliability weights
$w_{t,j,l}\propto\widehat\sigma_{t,j,l}^{-2}$, where
$\widehat\sigma_{t,j,l}^{2}$ estimates layer noise from within-window dispersion.
Appendix~\ref{app:kv_fusion} derives the fusion rule. This favors locally
consistent VLM depths without learned fusion parameters while matching the
proposer's smaller depth. Residual and feed-forward updates yield
$H_t^p=H_{t,L_p}^p$; we read out $v_\phi^p=W_{\mathrm{out}}^pH_t^p$ and take
one Euler step from $t_p=1$ to $0$:
\begin{equation}
\small
P_t
=
X_t^p-v_\phi^p,
\qquad
\mathcal{L}_{\mathrm{prop}}
=\frac{1}{H}\sum_k\|P_{t,k}-a_{t,k}^{*}\|_2^2 .
\label{eq:proposal_readout}
\end{equation}
This endpoint objective requires no additional parameterization or iterative proposal refinement. At inference, we simply remove the training perturbation and set $X_t^p=A_t^{\mathrm{mom}}$.

\vspace{-2pt}
\subsection{Asymmetric Desired-Future Modeling and Proposal Calibration}
\label{sec:world_expert}
\vspace{-2pt}

A single World Expert jointly reasons over the proposal and VLM semantics to capture proposal-conditioned prospective information. It is a Transformer without flow-time conditioning whose shared queries support asymmetric future-prediction and proposal-calibration readouts.

\textbf{V-JEPA future latent target.}
Rather than reconstructing pixels or imposing object slots, we supervise
prospective reasoning in video representation space. We form the future clip
$\mathcal{V}_t^{\rightarrow}
=(O_{t+1},O_{t+2},\ldots,O_{t+H})$ and
use the frozen V-JEPA~2 encoder from V-JEPA~2-AC~\cite{assran2025vjepa2} as the
target space. With temporal tubelet size $2$, $N^f=H/2$ is its temporal
resolution. We spatially pool the encoder tokens into $N_z$ groups within each
temporal tubelet:
\begin{equation}
\small
Z_t^{*}
=
\mathcal{P}_s\!\left(
\sg\!\left(E_{\mathrm{VJ}}(\mathcal{V}_t^{\rightarrow})\right)
\right)
=
\{z_{t,\tau,j}^{*}\}_{\tau,j}
\in\mathbb{R}^{N^f\times N_z\times d_f},
\label{eq:vjepa_future_target}
\end{equation}
where $\mathcal{P}_s$ performs spatial pooling and feature-wise normalization.
This removes the frozen encoder's arbitrary scale and balances future prediction
against the remaining objectives. The target captures physical evolution
without reconstructing task-irrelevant texture.

\textbf{Proposal-aware world reasoning.}
The World Expert initializes learnable spatiotemporal queries
$H_{t,0}^{w}=\{q_j^{w}+e_k^{\mathrm{time}}\}_{k,j}
\in\mathbb{R}^{H\times N_z\times d_w}$ at the full action-horizon
resolution and embeds the detached proposal as $H$ condition tokens,
$H_{t,0}^{c}=W_{\mathrm{in}}^{c}\sg(P_t)$. Only the final $P_t$ is shared.
Writing $S_{t,l}=[H_{t,l}^{w};H_{t,l}^{c}]$, each layer jointly attends to the
VLM cache and its own state:
\begin{equation}
\small
\overline S_{t,l}
=
\Attn\!\left(
Q_l^{w}(S_{t,l-1}),
\left[
K_{t,l}^{v};
K_{t,l}^{s}
\right],
\left[
V_{t,l}^{v};
V_{t,l}^{s}
\right]
\right),
\label{eq:world_joint_attention}
\end{equation}
where $K_{t,l}^{s},V_{t,l}^{s}$ project $S_{t,l-1}$, followed by standard
Transformer updates. VLM features provide task and scene semantics, while
proposal tokens inject a concrete motion tendency. Retaining only the
$H\times N_z$ world-query outputs gives $H_t^w=\{h_{t,k,j}^w\}_{k,j}$, a shared
\emph{proposal-conditioned prospective representation} for calibration.
Joint attention exposes proposal--scene compatibility per horizon step rather
than compressing the motion hypothesis into one global token.

\textbf{Asymmetric future and calibration readouts.}
For future prediction, neighboring world-token groups are pooled to match the
V-JEPA resolution,
$\overline h_{t,\tau,j}^{w}
=\operatorname{Pool}_{t}(h_{t,2\tau-1,j}^{w},h_{t,2\tau,j}^{w})$,
and a linear head maps $\overline h_{t,\tau,j}^{w}$ to
$\widehat z_{t,\tau,j}$. Fixed token ordering avoids matching, giving
$\mathcal{L}_{\mathrm{future}}
=\frac{1}{N^fN_zd_f}\sum_{\tau,j}\|\widehat z_{t,\tau,j}-z_{t,\tau,j}^{*}\|_2^2$.
The target is the demonstrated future, not the counterfactual outcome of
executing $P_t$, so the proposal remains a \emph{soft motion prior}.

For calibration, we pool latent tokens into one horizon-aligned feature,
$g_{t,k}^{w}=\frac{1}{N_z}\sum_j h_{t,k,j}^{w}$, and denote
$G_t^w=(g_{t,1}^w,\ldots,g_{t,H}^w)$. The future head decodes \emph{what
task-consistent evolution should look like}; calibration decodes \emph{how the
proposal must be loosened and reshaped} from the same prospective representation. Both calibration targets use residual $A_t^{*}-P_t$, making proposal information necessary even though the future target itself remains task-consistent. This asymmetry also prevents an inaccurate proposal from redefining the desired future merely to justify itself.

\begin{figure}[t]
\vspace{-4mm}
\centering
\includegraphics[width=1.0\textwidth]{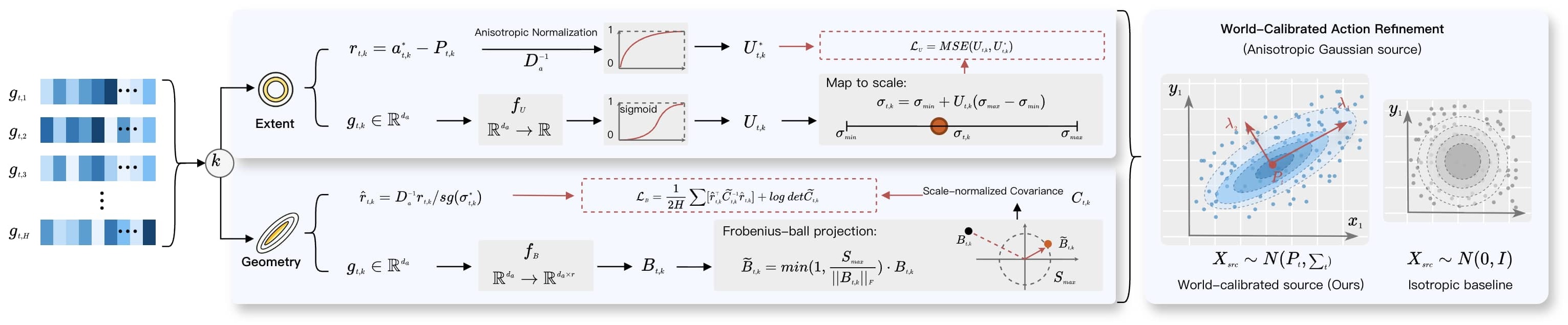}
\vspace{-6mm}
\caption{
\textbf{World-calibrated source construction.}
Proposal-conditioned future features yield per-step refinement extent and
low-rank geometry, defining an anisotropic source for action refinement.
}
\label{fig:detail}
\end{figure}

\vspace{-2pt}
\subsection{World-Calibrated Action Refinement}
\label{sec:action_refinement}
\vspace{-2pt}

The proposal-aligned $G_t^w$ defines a source with scalar \emph{extent} and
low-rank \emph{geometry}.

\textbf{Refinement extent.}
Let $r_{t,k}=a_{t,k}^{*}-P_{t,k}$, with the diagonal action-scale matrix $D_a$
(See Appendix~\ref{app:derivations}). We infer correction extent via
proposal--future compatibility:
\begin{equation}
\small
\begin{aligned}
e_{t,k}&=\frac{\|D_a^{-1}r_{t,k}\|_2}{\sqrt{d_a}},
&U_{t,k}^{*}&=1-\exp(-e_{t,k}/\tau_u),\\
U_{t,k}&=\operatorname{sigmoid}(f_u(g_{t,k}^{w})),
&\sigma_{t,k}^{(*)}&=\sigma_{\min}+U_{t,k}^{(*)}
(\sigma_{\max}-\sigma_{\min}).
\end{aligned}
\label{eq:extent}
\end{equation}
Here $\sigma_{t,k}^{*}$ uses $U_{t,k}^{*}$, while $\sigma_{t,k}$ uses
$U_{t,k}$. MSE $\mathcal{L}_U$ supervises $U_{t,k}$ against $U_{t,k}^{*}$.
Compatible proposals induce concentrated sources; incompatible ones open more
room for refinement.

\textbf{Refinement geometry.}
Magnitude alone misses dependencies among translation, rotation, and gripper
dimensions. From the same world feature, we predict a low-rank factor
$B_{t,k}=f_B(g_{t,k}^{w})\in\mathbb{R}^{d_a\times r}$, with $r\ll d_a$.
It defines the scale-normalized covariance at each proposal horizon step:
\begin{equation}
\small
C_{t,k}
=
\sigma_{t,k}^{2} d_a
\left(I+B_{t,k}B_{t,k}^{\top}\right)
/
\operatorname{tr}\!\left(I+B_{t,k}B_{t,k}^{\top}\right).
\label{eq:calibrated_covariance}
\end{equation}
To reduce magnitude dependence when learning refinement geometry, we normalize
residuals by the detached target extent and score them under the scale-normalized
$\widetilde C_{t,k}=C_{t,k}/\sigma_{t,k}^{2}$:
\begin{equation}
\small
\widehat r_{t,k}
=
\frac{D_a^{-1}r_{t,k}}{\sg(\sigma^{*}_{t,k})},
\qquad
\mathcal{L}_B
=
\frac{1}{2H}\sum_k
\left[
\widehat r_{t,k}^{\top}\widetilde C_{t,k}^{-1}\widehat r_{t,k}
+
\log\det\widetilde C_{t,k}
\right].
\label{eq:geometry_loss}
\end{equation}
Target normalization reduces residual-scale variation, while trace normalization
prevents $B_{t,k}$ from changing the average variance across action dimensions. Thus,
$\mathcal{L}_{U}$ primarily controls refinement magnitude and
$\mathcal{L}_{B}$ its geometry. Trace normalization removes scale but not
single-sample shape degeneracy, so we project
$B_{t,k}$ onto
$\|B_{t,k}\|_{F}^{2}\le S_{\max}$. This enforces
$\kappa(\widetilde C_{t,k})\le 1+S_{\max}$ while preserving
$\operatorname{tr}C_{t,k}=d_a\sigma_{t,k}^{2}$. We use $S_{\max}=5$;
Appendix~\ref{app:calibration} gives the derivation and $O(d_ar^2)$ evaluation.
Consequently, $\sigma_{t,k}$ fixes average variance while $B_{t,k}$ redistributes
it toward coupled residual directions without silently changing the overall refinement extent.

With action-space covariance $C_{t,k}^a=D_aC_{t,k}D_a^\top$, we sample the source as
\begin{equation}
\small
X_{t,\mathrm{src}}^{a}
=
P_t
+
L_t\epsilon_a,
~~~
\epsilon_a\sim\mathcal{N}(0,I),
~~~
L_tL_t^{\top}
=
\BlockDiag
\left(
C_{t,1}^a,\ldots,C_{t,H}^a
\right).
\label{eq:calibrated_source}
\end{equation}
Appendix~\ref{app:calibration} details sampling and covariance construction.

\textbf{Action flow refinement.}
The calibrated covariance defines preferred correction axes but not the signed
transport direction, which the Action Expert supplies. We preserve the base VLA's
flow parameterization, where $s=1$ is the source and $s=0$ the data, replacing
only its Gaussian source by $X_{t,\mathrm{src}}^{a}$. For base-model flow time
$s$ and velocity $v_\theta^A=W_vH_{t,s,L_a}^{a}$, the path and objective are
\begin{equation}
\small
X_{t,s}^{a}=sX_{t,\mathrm{src}}^{a}+(1-s)A_t^{*},~~~~
v_t^{*}=X_{t,\mathrm{src}}^{a}-A_t^{*},~~~~
\mathcal{L}_{\mathrm{flow}}
=\mathbb{E}_{t,s,\epsilon_p,\epsilon_a}
\left[\left\|v_\theta^A-v_t^{*}\right\|_2^2\right].
\label{eq:action_flow}
\end{equation}
The Action Expert uses $X_{t,s}^{a}$ as its action state, reads shared VLM
context, and attends to $H_t^w$ for proposal-aware prospective guidance.
Its velocity therefore receives both desired future evolution and proposal
compatibility, with attention and integration detailed in
Appendix~\ref{app:action_expert}.

At inference, we construct $P_t$ and $H_t^w$, predict $U_t$ and $B_t$, sample
$X_{t,\mathrm{src}}^{a}$, and integrate from $s=1$ to $0$.
Starting closer to the action manifold shortens the required transport,
allowing a coarser integration grid while reducing both sampling cost and
accumulated integration error.

\textbf{Training objective.}
The complete model is trained end-to-end with
\begin{equation}
\small
\mathcal{L}
=
\lambda_p\mathcal{L}_{\mathrm{prop}}
+\lambda_f\mathcal{L}_{\mathrm{future}}
+\lambda_u\mathcal{L}_{U}
+\lambda_b\mathcal{L}_{B}
+\lambda_a\mathcal{L}_{\mathrm{flow}} .
\label{eq:total_loss}
\end{equation}
These terms learn a useful proposal, anchor world reasoning in V-JEPA space,
calibrate proposal-specific extent and geometry, and transport samples toward
executable actions.

\begin{table}[t]
\vspace{-4mm}

\centering


\caption{\textbf{Results on LIBERO and LIBERO-Plus.}
We report performance on the four LIBERO task suites and generalization under
seven perturbation dimensions in LIBERO-Plus.}
\label{tab:libero_libero_plus}

\setlength{\tabcolsep}{4.8pt}

\resizebox{\textwidth}{!}{%
\begin{tabular}{lccccc||cccccccc}
\toprule

\raisebox{-0.45\height}[0pt][0pt]{%
    \parbox[c][0.9cm][c]{1.55cm}{\raggedright\textbf{Method}}%
}
& \multicolumn{5}{c||}{\textbf{LIBERO}}
& \multicolumn{8}{c}{\textbf{LIBERO-Plus}} \\

\cmidrule(lr){2-6}
\cmidrule(lr){7-14}

&
Spatial
& Object
& Goal
& Long
& Avg.
& Camera
& Robot
& Language
& Light
& Background
& Noise
& Layout
& Total \\

\midrule

OpenVLA-OFT~\cite{oft}
& 97.6 & 98.4 & 97.9 & 94.5 & 97.1
& 56.4 & 31.9 & 79.5 & 88.7 & 93.3 & 75.8 & 74.2 & 69.6 \\

$\pi_{0.5}$~\cite{pi05}
& 98.8 & 98.2 & 98.0 & 92.4 & 96.9
& 59.7 & 65.5 & 75.3 & 87.0 & 82.4 & 72.1 & 80.3 & 73.6 \\

WorldVLA~\cite{cen2025worldvla}
& 87.6 & 96.2 & 83.4 & 60.0 & 81.8
& 0.1 & 27.9 & 41.6 & 43.7 & 17.1 & 10.9 & 38.0 & 25.0 \\

UniVLA~\cite{bu2025univla}
& 95.4 & 98.8 & 93.6 & 94.0 & 95.4
& 1.8 & 46.2 & 69.6 & 69.0 & 81.0 & 21.2 & 31.9 & 42.9 \\

VLA-Adapter~\cite{wang2026vla}
& 97.8 & 99.2 & 97.2 & 95.0 & 97.3
& 36.2 & 37.9 & 74.6 & 70.6 & 76.1 & 58.0 & 69.7 & 59.1 \\

DreamVLA~\cite{zhang2026dreamvla}
& 97.5 & 94.0 & 89.5 & 89.5 & 92.6
& 65.0 & 40.8 & 63.5 & 85.7 & 82.6 & 84.9 & 74.0 & 69.9 \\

VLA-JEPA~\cite{sun2026vlajepa}
& 96.2 & 99.6 & 97.2 & 95.8 & 97.2
& 63.3 & 67.1 & 85.4 & 95.6 & 93.6 & 66.3 & 85.1 & 79.5 \\

Fast-WAM~\cite{yuan2026fast}
& 98.2 & \textbf{100.0} & 98.2 & 95.2 & 97.6
& 34.0 & 55.1 & \textbf{88.9} & 90.0 & 44.9 & 33.3 & 73.4 & 59.0 \\

HoloBrain~\cite{lin2026holobrain}
& -- & -- & -- & -- & --
& 66.3 & 49.0 & 65.9 & 94.9 & 93.3 & 73.1 & 78.2 & 72.6 \\

\midrule

\rowcolor[HTML]{DDE7F4}
\method{}
& \textbf{99.4}
& 99.6
& \textbf{98.4}
& \textbf{96.0}
& \textbf{98.4}
& \textbf{76.1}
& \textbf{80.3}
& 87.4
& \textbf{97.8}
& \textbf{95.5}
& \textbf{90.2}
& \textbf{86.2}
& \textbf{86.8} \\

\bottomrule
\end{tabular}%
}

\par\medskip

\vspace{-1em}


\caption{\textbf{Results on RoboTwin 2.0 benchmark.}
All policies are trained on the Clean setting and evaluated under the
{Randomized (hard) zero-shot} setting, with 100 rollouts per task.
}
\label{tab:robotwin_random_subset}

\setlength{\tabcolsep}{3.5pt}

\resizebox{\textwidth}{!}{
\begin{tabular}{l|ccccccccccc}
\toprule

\parbox[c][1.25cm][c]{1.35cm}{\raggedright\textbf{Method}}
& \parbox[c][1.25cm][c]{1.35cm}{\centering Put\\Bottles\\Dustbin}
& \parbox[c][1.25cm][c]{1.35cm}{\centering Place\\Dual\\Shoes}
& \parbox[c][1.25cm][c]{1.35cm}{\centering Pick\\Diverse\\Bottles}
& \parbox[c][1.25cm][c]{1.35cm}{\centering Place\\Burger\\Fries}
& \parbox[c][1.25cm][c]{1.35cm}{\centering Open\\Laptop}
& \parbox[c][1.25cm][c]{1.35cm}{\centering Blocks\\Ranking\\RGB}
& \parbox[c][1.25cm][c]{1.35cm}{\centering Pick\\Dual\\Bottles}
& \parbox[c][1.25cm][c]{1.35cm}{\centering Dump\\Bin\\Bigbin}
& \parbox[c][1.25cm][c]{1.35cm}{\centering Stamp\\Seal}
& \parbox[c][1.25cm][c]{1.35cm}{\centering Place\\Bread\\Basket}
& \parbox[c][1.25cm][c]{0.75cm}{\centering Avg.} \\

\midrule


$\pi_0$~\cite{pi0}
& 13\% & 0\% & 6\% & 4\% & 46\%
& 5\% & 12\% & 24\% & 4\% & 4\%
& 11.8\% \\

$\pi_{0.5}$~\cite{pi05}
& 9\% & 7\% & 3\% & 70\% & 35\%
& 35\% & 6\% & 42\% & 23\% & 56\%
& 28.6\% \\

X-VLA~\cite{zheng2026x}
& 20\% & 12\% & 30\% & 35\% & 16\%
& 4\% & 51\% & 41\% & 2\% & 52\%
& 26.3\% \\

Spatial Forcing~\cite{li2026spatial}
& 41\% & 2\% & 38\% & 71\% & 15\%
& 7\% & 47\% & 70\% & 2\% & 53\%
& 34.6\% \\

X-WAM~\cite{guo2026unified}
& 21\% & 7\% & 26\% & 36\% & 6\%
& 36\% & 55\% & 40\% & 16\% & 40\%
& 28.3\% \\

Abot-M0~\cite{yang2026abot}
& 26\% & 15\% & 7\% & 55\% & 64\%
& 23\% & 19\% & 68\% & 19\% & 47\%
& 34.3\% \\

EventVLA~\cite{yang2026eventvla}
& 31\% & 1\% & 9\% & 7\% & 23\%
& 6\% & 8\% & 64\% & 3\% & 15\%
& 16.7\% \\

Xiaomi Robotics-0~\cite{cai2026xiaomi}
& 21\% & 9\% & 12\% & 27\% & 24\%
& 8\% & 28\% & 49\% & 3\% & 23\%
& 20.4\% \\

FastWAM~\cite{yuan2026fast}
& 0\% & 0\% & 0\% & 0\% & 0\%
& 0\% & 1\% & 0\% & 0\% & 0\%
& 0.1\% \\

GalaxeaVLA~\cite{liu2026g05autoregressivestreamrobot}
& 3\% & 9\% & 11\% & 32\% & 13\%
& 1\% & 21\% & 40\% & 2\% & 32\%
& 16.4\% \\


starVLA~\cite{community2026starvla}
& 0\% & 2\% & 15\% & 11\% & 2\%
& 2\% & 20\% & 4\% & 0\% & 5\%
& 6.1\% \\

\midrule

\rowcolor[HTML]{DDE7F4}
\textbf{\method{}}
& \textbf{69\%}
& \textbf{39\%}
& \textbf{57\%}
& \textbf{84\%}
& \textbf{76\%}
& \textbf{46\%}
& \textbf{63\%}
& \textbf{78\%}
& \textbf{29\%}
& \textbf{62\%}
& \textbf{60.3\%} \\

\bottomrule
\end{tabular}
}

\end{table}

\vspace{-6pt}
\section{Experiments}
\label{sec:experiments}
\vspace{-3pt}

We evaluate \method{} through four questions: \textbf{(Q1)} Does the world-calibrated,
proposal-centered source outperform strong VLA baselines? \textbf{(Q2)} Are gains larger
when long-horizon or perturbed conditions weaken motion continuity? \textbf{(Q3)} Which
propose--calibrate--refine components drive the gains? \textbf{(Q4)} Does calibration
improve proposal refinement and generation efficiency?

\vspace{-2pt}
\subsection{Comparison on Simulation Benchmarks}
\label{sec:exp_simulation}
\vspace{-2pt}

This section addresses \textbf{Q1} and \textbf{Q2} on three complementary
benchmarks, testing whether a calibrated source improves strong VLA baselines
under saturation, distribution shift, and long-horizon bimanual control across
increasingly demanding manipulation conditions.

\textbf{LIBERO Benchmark.}
LIBERO~\cite{liu2023libero} spans spatial, object, goal, and long-horizon
variation; following $\pi_{0.5}$~\cite{pi05}, we report 500 rollouts per suite.
Tab.~\ref{tab:libero_libero_plus} shows that \method{} achieves 98.4\% average
success, 1.5 points above its backbone. While single-stage suites are near
saturation, the margin grows to 3.6 points on Long, supporting calibrated
source expansion when stage transitions sharply break local motion continuity.

\begin{figure}[t]
    \vspace{-4mm}
    \includegraphics[width=1.0\textwidth]{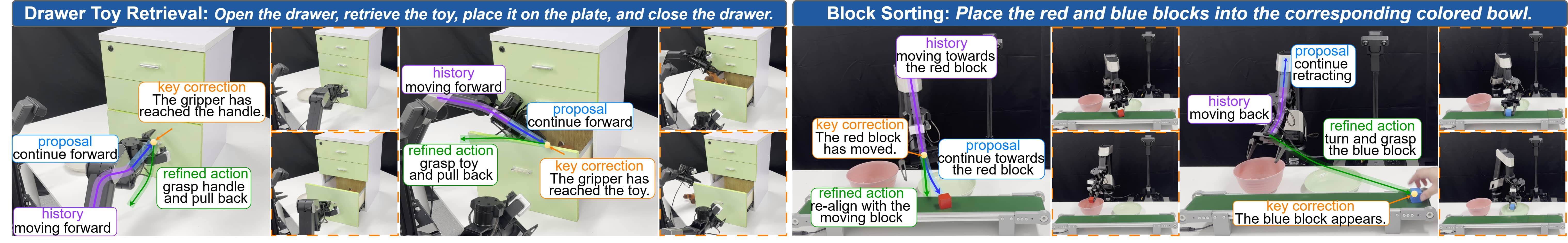}
    \vspace{-6mm}
    \caption{
    \textbf{Closed-loop proposal refinement on real robots.}
    Drawer Toy Retrieval (left) and Block Sorting (right). Enlarged frames overlay
    history (purple), proposal (blue), and refined action (green); orange callouts
    mark changes requiring departure from local motion continuity.
    }
    \label{fig:vis-real}
    \vspace{-2mm}
\end{figure}

\begin{table}[t]
\vspace{-2mm}

\centering
\caption{
\textbf{Real-world manipulation results.}
Results are reported over the same six tasks on both  on GALAXEA R1 Lite~\cite{galaxea} and AgileX Cobot Magic~\cite{aloha} platforms.
The superscript $\dagger$ denotes our reproduced baselines.
}
\label{tab:real_world}


{
\setlength{\fboxsep}{1pt}%
\colorbox[HTML]{EFEFEF}{%
    \makebox[\dimexpr\textwidth-2\fboxsep\relax][c]{%
        \fontsize{6.5pt}{7.5pt}\selectfont
        \textbf{Galaxea R1 Lite Platform}%
    }%
}%
}
\par\vspace{-0.4pt}


\setlength{\tabcolsep}{3.0pt}

\resizebox{\textwidth}{!}{%
\begin{tabular}{l|ccccc|ccc|ccccc}
\toprule

\raisebox{-0.45\height}[0pt][0pt]{%
    \parbox[c][1.05cm][c]{1.65cm}{%
        \raggedright\textbf{Method}%
    }%
}
& \multicolumn{5}{c|}{\textbf{Block Sorting}}
& \multicolumn{3}{c|}{\textbf{Pot Lid Retrieval}}
& \multicolumn{5}{c}{\textbf{Drawer Manipulation}} \\

\cmidrule(lr){2-6}
\cmidrule(lr){7-9}
\cmidrule(lr){10-14}

& \shortstack{Pick\\Red Block}
& \shortstack{+Place Red\\in Bowl}
& \shortstack{+Pick\\Blue Block}
& \shortstack{+Place Blue\\in Bowl}
& \textbf{CSR}

& \shortstack{Remove\\Lid}
& \shortstack{+Retrieve\\Object to Plate}
& \textbf{CSR}

& \shortstack{Open\\Drawer}
& \shortstack{+Grasp\\Toy}
& \shortstack{+Place Toy\\Inside}
& \shortstack{+Close\\Drawer}
& \textbf{CSR} \\

\midrule

OpenVLA-OFT~\cite{oft}$^{\dagger}$
& 19{\fontsize{6pt}{7pt}\selectfont /25}
& 17{\fontsize{6pt}{7pt}\selectfont /25}
& 13{\fontsize{6pt}{7pt}\selectfont /25}
& 12{\fontsize{6pt}{7pt}\selectfont /25}
& 61.0\%
& 21{\fontsize{6pt}{7pt}\selectfont /25}
& 17{\fontsize{6pt}{7pt}\selectfont /25}
& 76.0\%
& 17{\fontsize{6pt}{7pt}\selectfont /25}
& 15{\fontsize{6pt}{7pt}\selectfont /25}
& 12{\fontsize{6pt}{7pt}\selectfont /25}
& 12{\fontsize{6pt}{7pt}\selectfont /25}
& 56.0\% \\

$\pi_0$~\cite{pi0}$^{\dagger}$
& 20{\fontsize{6pt}{7pt}\selectfont /25}
& 18{\fontsize{6pt}{7pt}\selectfont /25}
& 14{\fontsize{6pt}{7pt}\selectfont /25}
& 14{\fontsize{6pt}{7pt}\selectfont /25}
& 66.0\%
& 21{\fontsize{6pt}{7pt}\selectfont /25}
& 16{\fontsize{6pt}{7pt}\selectfont /25}
& 74.0\%
& 18{\fontsize{6pt}{7pt}\selectfont /25}
& 17{\fontsize{6pt}{7pt}\selectfont /25}
& 13{\fontsize{6pt}{7pt}\selectfont /25}
& 13{\fontsize{6pt}{7pt}\selectfont /25}
& 61.0\% \\

$\pi_{0.5}$~\cite{pi05}$^{\dagger}$
& 23{\fontsize{6pt}{7pt}\selectfont /25}
& 22{\fontsize{6pt}{7pt}\selectfont /25}
& 19{\fontsize{6pt}{7pt}\selectfont /25}
& 17{\fontsize{6pt}{7pt}\selectfont /25}
& 81.0\%
& 22{\fontsize{6pt}{7pt}\selectfont /25}
& 20{\fontsize{6pt}{7pt}\selectfont /25}
& 84.0\%
& 21{\fontsize{6pt}{7pt}\selectfont /25}
& 21{\fontsize{6pt}{7pt}\selectfont /25}
& 18{\fontsize{6pt}{7pt}\selectfont /25}
& 16{\fontsize{6pt}{7pt}\selectfont /25}
& 76.0\% \\

\rowcolor[HTML]{DDE7F4}
\textbf{\method{}}
& \textbf{23{\fontsize{6pt}{7pt}\selectfont /25}}
& \textbf{23{\fontsize{6pt}{7pt}\selectfont /25}}
& \textbf{21{\fontsize{6pt}{7pt}\selectfont /25}}
& \textbf{20{\fontsize{6pt}{7pt}\selectfont /25}}
& \textbf{87.0\%}
& \textbf{24{\fontsize{6pt}{7pt}\selectfont /25}}
& \textbf{21{\fontsize{6pt}{7pt}\selectfont /25}}
& \textbf{90.0\%}
& \textbf{23{\fontsize{6pt}{7pt}\selectfont /25}}
& \textbf{22{\fontsize{6pt}{7pt}\selectfont /25}}
& \textbf{20{\fontsize{6pt}{7pt}\selectfont /25}}
& \textbf{19{\fontsize{6pt}{7pt}\selectfont /25}}
& \textbf{84.0\%} \\

\midrule
\end{tabular}%
}

\par\vspace{-2pt}


\setlength{\tabcolsep}{3.1pt}

\resizebox{\textwidth}{!}{%
\begin{tabular}{l|ccccc|ccc|cccc|cc}

\raisebox{-0.45\height}[0pt][0pt]{%
    \parbox[c][1.05cm][c]{1.65cm}{%
        \raggedright\textbf{Method}%
    }%
}
& \multicolumn{5}{c|}{\textbf{Drawer Toy Retrieval}}
& \multicolumn{3}{c|}{\textbf{Table Wiping}}
& \multicolumn{4}{c|}{\textbf{Color Sorting}}
& \multicolumn{2}{c}{\textbf{Overall}} \\

\cmidrule(lr){2-6}
\cmidrule(lr){7-9}
\cmidrule(lr){10-13}
\cmidrule(lr){14-15}

& \shortstack{Open\\Drawer}
& \shortstack{+Retrieve\\Toy}
& \shortstack{+Place Toy\\on Plate}
& \shortstack{+Close\\Drawer}
& \textbf{CSR}

& \shortstack{Grasp\\Cloth}
& \shortstack{+Wipe\\Table}
& \textbf{CSR}

& \shortstack{Red\\Block$\rightarrow$Bin}
& \shortstack{+Blue\\Block$\rightarrow$Bin}
& \shortstack{+Yellow\\Block$\rightarrow$Bin}
& \textbf{CSR}

& \textbf{TSR}
& \textbf{CSR} \\

\midrule

OpenVLA-OFT~\cite{oft}$^{\dagger}$
& 15{\fontsize{6pt}{7pt}\selectfont /25}
& 9{\fontsize{6pt}{7pt}\selectfont /25}
& 8{\fontsize{6pt}{7pt}\selectfont /25}
& 7{\fontsize{6pt}{7pt}\selectfont /25}
& 39.0\%
& 19{\fontsize{6pt}{7pt}\selectfont /25}
& 12{\fontsize{6pt}{7pt}\selectfont /25}
& 62.0\%
& 22{\fontsize{6pt}{7pt}\selectfont /25}
& 16{\fontsize{6pt}{7pt}\selectfont /25}
& 9{\fontsize{6pt}{7pt}\selectfont /25}
& 62.7\%
& 46.0\%
& 59.4\% \\

$\pi_0$~\cite{pi0}$^{\dagger}$
& 19{\fontsize{6pt}{7pt}\selectfont /25}
& 14{\fontsize{6pt}{7pt}\selectfont /25}
& 14{\fontsize{6pt}{7pt}\selectfont /25}
& 12{\fontsize{6pt}{7pt}\selectfont /25}
& 59.0\%
& 22{\fontsize{6pt}{7pt}\selectfont /25}
& 16{\fontsize{6pt}{7pt}\selectfont /25}
& 76.0\%
& 20{\fontsize{6pt}{7pt}\selectfont /25}
& 16{\fontsize{6pt}{7pt}\selectfont /25}
& 13{\fontsize{6pt}{7pt}\selectfont /25}
& 65.3\%
& 56.0\%
& 66.9\% \\

$\pi_{0.5}$~\cite{pi05}$^{\dagger}$
& 22{\fontsize{6pt}{7pt}\selectfont /25}
& 18{\fontsize{6pt}{7pt}\selectfont /25}
& 17{\fontsize{6pt}{7pt}\selectfont /25}
& 16{\fontsize{6pt}{7pt}\selectfont /25}
& 73.0\%
& 24{\fontsize{6pt}{7pt}\selectfont /25}
& 20{\fontsize{6pt}{7pt}\selectfont /25}
& 88.0\%
& 23{\fontsize{6pt}{7pt}\selectfont /25}
& 18{\fontsize{6pt}{7pt}\selectfont /25}
& 16{\fontsize{6pt}{7pt}\selectfont /25}
& 76.0\%
& 70.0\%
& 79.7\% \\

\rowcolor[HTML]{DDE7F4}
\textbf{\method{}}
& \textbf{22{\fontsize{6pt}{7pt}\selectfont /25}}
& \textbf{20{\fontsize{6pt}{7pt}\selectfont /25}}
& \textbf{19{\fontsize{6pt}{7pt}\selectfont /25}}
& \textbf{19{\fontsize{6pt}{7pt}\selectfont /25}}
& \textbf{80.0\%}
& \textbf{24{\fontsize{6pt}{7pt}\selectfont /25}}
& \textbf{21{\fontsize{6pt}{7pt}\selectfont /25}}
& \textbf{90.0\%}
& \textbf{24{\fontsize{6pt}{7pt}\selectfont /25}}
& \textbf{22{\fontsize{6pt}{7pt}\selectfont /25}}
& \textbf{18{\fontsize{6pt}{7pt}\selectfont /25}}
& \textbf{85.3\%}
& \textbf{78.7\%}
& \textbf{86.1\%} \\

\bottomrule
\end{tabular}%
}

\par\vspace{2pt}


{
\setlength{\fboxsep}{1pt}%
\colorbox[HTML]{EFEFEF}{%
    \makebox[\dimexpr\textwidth-2\fboxsep\relax][c]{%
        \fontsize{6.5pt}{7.5pt}\selectfont
        \textbf{AgileX Cobot Magic Platform}%
    }%
}%
}
\par\vspace{-0.4pt}

\setlength{\tabcolsep}{3.8pt}

{%
\newcommand{\metricpair}[2]{%
    \makebox[0.85cm][c]{#1}%
    \hspace{0.30cm}%
    \makebox[0.85cm][c]{#2}%
}

\resizebox{\textwidth}{!}{%
\begin{tabular}{l|c|c|c|c|c|c|c}
\toprule

\textbf{Method}
& \textbf{Block Sorting}
& \textbf{Pot Lid Retrieval}
& \textbf{Drawer Manipulation}
& \textbf{Drawer Toy Retrieval}
& \textbf{Table Wiping}
& \textbf{Color Sorting}
& \textbf{Average}
\\

& \metricpair{\textbf{TSR}}{\textbf{CSR}}
& \metricpair{\textbf{TSR}}{\textbf{CSR}}
& \metricpair{\textbf{TSR}}{\textbf{CSR}}
& \metricpair{\textbf{TSR}}{\textbf{CSR}}
& \metricpair{\textbf{TSR}}{\textbf{CSR}}
& \metricpair{\textbf{TSR}}{\textbf{CSR}}
& \metricpair{\textbf{TSR}}{\textbf{CSR}}
\\

\midrule

OpenVLA-OFT~\cite{oft}$^{\dagger}$
& \metricpair{44.0\%}{55.0\%}
& \metricpair{60.0\%}{70.0\%}
& \metricpair{40.0\%}{52.0\%}
& \metricpair{28.0\%}{37.0\%}
& \metricpair{48.0\%}{62.0\%}
& \metricpair{36.0\%}{57.3\%}
& \metricpair{42.7\%}{55.6\%}
\\

$\pi_0$~\cite{pi0}$^{\dagger}$
& \metricpair{52.0\%}{63.0\%}
& \metricpair{56.0\%}{68.0\%}
& \metricpair{48.0\%}{59.0\%}
& \metricpair{44.0\%}{54.0\%}
& \metricpair{56.0\%}{68.0\%}
& \metricpair{48.0\%}{61.3\%}
& \metricpair{50.7\%}{62.2\%}
\\

$\pi_{0.5}$~\cite{pi05}$^{\dagger}$
& \metricpair{64.0\%}{73.0\%}
& \metricpair{72.0\%}{78.0\%}
& \metricpair{60.0\%}{70.0\%}
& \metricpair{56.0\%}{65.0\%}
& \metricpair{72.0\%}{82.0\%}
& \metricpair{64.0\%}{76.0\%}
& \metricpair{64.7\%}{74.0\%}
\\

\rowcolor[HTML]{DDE7F4}
\textbf{\method{}}
& \metricpair{\textbf{76.0\%}}{\textbf{82.0\%}}
& \metricpair{\textbf{80.0\%}}{\textbf{86.0\%}}
& \metricpair{\textbf{72.0\%}}{\textbf{80.0\%}}
& \metricpair{\textbf{68.0\%}}{\textbf{75.0\%}}
& \metricpair{\textbf{80.0\%}}{\textbf{88.0\%}}
& \metricpair{\textbf{72.0\%}}{\textbf{82.7\%}}
& \metricpair{\textbf{74.7\%}}{\textbf{82.3\%}}
\\

\bottomrule
\end{tabular}%
}
}

\end{table}

\textbf{LIBERO-Plus Benchmark.}
Without retraining, we test seven LIBERO-Plus perturbations~\cite{fei2025libero}
under the Total protocol. Tab.~\ref{tab:libero_libero_plus} shows that
\method{} reaches 86.8\%, 13.2 points above $\pi_{0.5}$ and 7.3 above the strongest
competing average. It improves every category, with the largest gains on Noise,
Camera, and Robot shifts (18.1/16.4/14.8 points), showing that
calibration re-anchors motion proposal to altered visual and geometric conditions.

\textbf{RoboTwin 2.0 Benchmark.}
We train one policy on all 50 Aloha-AgileX tasks~\cite{chen2025robotwin} and
evaluate 100 zero-shot \textit{Randomized (Hard)} rollouts per task.
Tab.~\ref{tab:robotwin_random_subset} reports the ten largest margins; full
results appear in Appendix~\ref{app:robotwin_full}. \method{} averages 60.3\%,
31.7 points above $\pi_{0.5}$ and 25.7 above the strongest competing average.
Gains across placement, bimanual coordination, and object-dependent grasping
show robustness across randomized scenes with unreliable local motion.

\vspace{-2pt}
\subsection{Comparison on Real-World Tasks}
\label{sec:exp_real_world}
\vspace{-2pt}

This section further addresses \textbf{Q1} and \textbf{Q2} on six tasks across
GALAXEA R1 Lite and AgileX Cobot Magic, with 25 trials each. TSR requires full
completion, while CSR retains checkpoint progress; Appendix~
\ref{app:real_metrics} provides metric details. All baselines use the same setup,
enabling direct comparison across methods and embodiments under identical protocols.

\textbf{Closed-loop correction on real robots.}
Fig.~\ref{fig:vis-real} shows selective refinement over complete rollouts. Forward
proposals become grasp-and-pull motions at contact in Drawer Toy Retrieval, while
Block Sorting realigns to displaced or newly relevant blocks. Proposals preserve
stable phases, and calibration handles contact and target transitions throughout
execution.

\textbf{Results on both robot platforms.}
Tab.~\ref{tab:real_world} shows consistent gains over $\pi_{0.5}$: overall
TSR/CSR rises by 8.7/6.4 points on GALAXEA and 10.0/8.3 on AgileX, with
12-point TSR gains on three AgileX tasks. The CSR--TSR gap narrows by 2.3/1.7
points, showing that intermediate progress more often survives task completion
across both physical embodiments and task settings. The consistent gains across
tasks further support the robustness of the refinement mechanism.

\vspace{-2pt}
\subsection{Ablation Study}
\label{sec:ablation}
\vspace{-2pt}

\newcommand{\cmark}{\ensuremath{\checkmark}}
\newcommand{\xmark}{\ensuremath{\times}}

\begin{table}[t]
\vspace{-6mm}

\centering
\scriptsize


\begin{minipage}[t]{0.485\linewidth}
\vspace{0pt}

\caption{
\textbf{Motion proposal ablation.}
Success rates (\%) under different source initializations for action generation,
from generic noise to the scene-aware motion proposal.
}
\label{tab:proposal_ablation}

\centering
\setlength{\tabcolsep}{1.8pt}
\renewcommand{\arraystretch}{1.28}

\begin{tabular*}{\linewidth}{
@{\extracolsep{\fill}}
lcccccc
@{}
}
\toprule

\multirow{2}{*}{\textbf{Initialization}}
&
\multicolumn{5}{c}{\textbf{LIBERO}}
&
\textbf{Real-World}
\\[-1.5pt]
\cmidrule(lr){2-6}
\cmidrule(lr){7-7}

&
\textbf{Spa.}
&
\textbf{Obj.}
&
\textbf{Goal}
&
\textbf{Long}
&
\textbf{Avg.}
&
\textbf{SR}
\\[-1.5pt]
\midrule

Gaussian $\mathcal{N}(0,I)$
& 98.8 & 98.6 & 97.8 & 93.6 & 97.2 & 72.0
\\

Last Action Chunk
& 98.4 & 98.2 & 97.4 & 92.8 & 96.7 & 70.7
\\

Motion Extrapolation
& 99.0 & 99.2 & 98.0 & 94.8 & 97.8 & 75.3
\\

\rowcolor[HTML]{DDE7F4}
\textbf{Scene-aware Proposal}
& 99.4
& 99.6
& 98.4
& 96.0
& \textbf{98.4}
& \textbf{78.7}
\\

\bottomrule
\end{tabular*}

\end{minipage}%
\hfill
\begin{minipage}[t]{0.485\linewidth}
\vspace{0pt}

\caption{
\textbf{World calibration ablation.}
We report success rates (\%) while ablating the World Expert (W),
proposal conditioning (P), future supervision (F), and source calibration (C).
}
\label{tab:world_ablation}

\centering
\setlength{\tabcolsep}{1.05pt}
\renewcommand{\arraystretch}{1.16}

\begin{tabular*}{\linewidth}{
@{\extracolsep{\fill}}
cccc|cccccc
@{}
}
\toprule

\multicolumn{4}{c|}{\textbf{Components}}
&
\multicolumn{5}{c}{\textbf{LIBERO}}
&
\textbf{Real-World}
\\[-1.5pt]
\cmidrule(lr){1-4}
\cmidrule(lr){5-9}
\cmidrule(lr){10-10}

\textbf{W}
&
\textbf{P}
&
\textbf{F}
&
\textbf{C}
&
\textbf{Spa.}
&
\textbf{Obj.}
&
\textbf{Goal}
&
\textbf{Long}
&
\textbf{Avg.}
&
\textbf{SR}
\\[-1.5pt]
\midrule

\cmark & \cmark & \xmark & \cmark
& 99.0 & 99.2 & 98.0 & 93.6 & 97.5 & 74.7
\\

\cmark & \xmark & \cmark & \cmark
& 98.8 & 99.0 & 97.6 & 92.8 & 97.1 & 73.3
\\

\cmark & \cmark & \cmark & \xmark
& 99.2 & 99.4 & 98.2 & 94.4 & 97.8 & 76.0
\\

\xmark & -- & -- & --
& 98.6 & 98.4 & 97.2 & 91.6 & 96.5 & 70.7
\\

\rowcolor[HTML]{DDE7F4}
\cmark & \cmark & \cmark & \cmark
& 99.4
& 99.6
& 98.4
& 96.0
& \textbf{98.4}
& \textbf{78.7}
\\

\bottomrule
\end{tabular*}

\end{minipage}


\begin{minipage}[t]{0.485\linewidth}
\vspace{0pt}

\caption{
\textbf{Source calibration ablation.}
Success rates (\%) across source variants; U and B denote refinement extent and geometry.
}
\label{tab:source_ablation}

\centering
\setlength{\tabcolsep}{1.35pt}
\renewcommand{\arraystretch}{1.22}

\begin{tabular*}{\linewidth}{
@{\extracolsep{\fill}}
lcc|cccccc
@{}
}
\toprule

\multicolumn{3}{c|}{\textbf{Calibration}}
&
\multicolumn{5}{c}{\textbf{LIBERO}}
&
\textbf{Real-World}
\\[-1.5pt]
\cmidrule(lr){1-3}
\cmidrule(lr){4-8}
\cmidrule(lr){9-9}

\textbf{Source}
&
\textbf{U}
&
\textbf{B}
&
\textbf{Spa.}
&
\textbf{Obj.}
&
\textbf{Goal}
&
\textbf{Long}
&
\textbf{Avg.}
&
\textbf{SR}
\\[-1.5pt]
\midrule

Fixed Isotropic
& \xmark & \xmark
& 98.6 & 98.8 & 97.4 & 92.8 & 96.9 & 72.7
\\

Extent Only
& \cmark & \xmark
& 99.0 & 99.2 & 98.0 & 94.8 & 97.8 & 76.0
\\

Geometry Only
& \xmark & \cmark
& 98.8 & 99.0 & 97.6 & 93.6 & 97.3 & 74.7
\\

\rowcolor[HTML]{DDE7F4}
\textbf{\method{}}
& \cmark
& \cmark
& 99.4
& 99.6
& 98.4
& 96.0
& \textbf{98.4}
& \textbf{78.7}
\\

\bottomrule
\end{tabular*}

\end{minipage}%
\hfill
\begin{minipage}[t]{0.485\linewidth}
\vspace{0pt}

\caption{
\textbf{Inference efficiency.}
Latency, throughput, and success rate under default inference on LIBERO
(A800, upper rows) and real-world tasks (RTX 5090, lower rows).
}
\label{tab:efficiency}

\centering
\setlength{\tabcolsep}{2.8pt}
\renewcommand{\arraystretch}{1.22}

\begin{tabular*}{\linewidth}{
@{\extracolsep{\fill}}
lccc
@{}
}
\toprule

\textbf{Model}
&
\textbf{Latency$\downarrow$}
&
\textbf{Throughput$\uparrow$}
&
\textbf{Success Rate$\uparrow$}
\\[-1.5pt]
\midrule

$\pi_{0.5}$
& 121.26 ms
& 82.47 Hz
& 96.9\%
\\

\rowcolor[HTML]{DDE7F4}
\textbf{\method{}}
& \textbf{89.98 ms}
& \textbf{111.14 Hz}
& \textbf{98.4\%}
\\

\midrule

$\pi_{0.5}$
& 104.62 ms
& 477.92 Hz
& 70.0\%
\\

\rowcolor[HTML]{DDE7F4}
\textbf{\method{}}
& \textbf{87.52 ms}
& \textbf{571.30 Hz}
& \textbf{78.7\%}
\\

\bottomrule
\end{tabular*}

\end{minipage}
\vspace{-3mm}

\end{table}

This section addresses \textbf{Q3} through component ablations in simulation and the real world.

\textbf{Motion-informed source.}
Tab.~\ref{tab:proposal_ablation} separates motion continuity from scene
grounding. Reusing the last chunk is slightly worse than Gaussian noise, whereas
motion extrapolation gains 0.6/3.3 points on LIBERO/real-world tasks. Proposal
Expert grounding adds another 0.6/3.4 points, showing that recent motion is a
useful soft prior but needs scene grounding when local continuity is unreliable.

\textbf{World calibration.}
Tab.~\ref{tab:world_ablation} validates the asymmetric calibration interface.
Removing the World Expert causes the largest drop, by 1.9 points on LIBERO and
8.0 points in the real world. Removing proposal conditioning,
future supervision, or source calibration costs 1.3/5.4, 0.9/4.0, and
0.6/2.7 points, respectively. Thus, proposal-aware future reasoning and source
calibration are both necessary, especially when real-world motion continuity
becomes unreliable. The larger real-world penalties highlight their importance
under physical manipulation settings.

\textbf{Refinement extent and geometry.}
Tab.~\ref{tab:source_ablation} shows that the two calibration outputs are
complementary. Relative to a fixed isotropic proposal-centered source, adaptive
extent improves LIBERO/real-world success by 0.9/3.3 points, while low-rank
geometry contributes 0.4/2.0 points. Combining both raises the gains to
1.5/6.0 points. Extent controls correction magnitude, while geometry allocates
it across coupled action dimensions. Their complementary gains show that
effective refinement needs both adaptive magnitude and structured directionality,
especially across embodiments, real-world contact transitions, and long-horizon
manipulation settings in practice.

\vspace{-2pt}
\subsection{Qualitative Analysis}
\label{sec:analysis}
\vspace{-2pt}

This section addresses \textbf{Q4} by examining whether the learned proposals
and calibrations are interpretable and whether proposal-centered transport
improves efficiency during action generation.

\suppressfloats[t]

\begin{figure}[t]
\vspace{-4mm}
\centering
\begin{minipage}[t]{0.661525\textwidth}
\vspace{0pt}
\centering
\includegraphics[width=\linewidth]{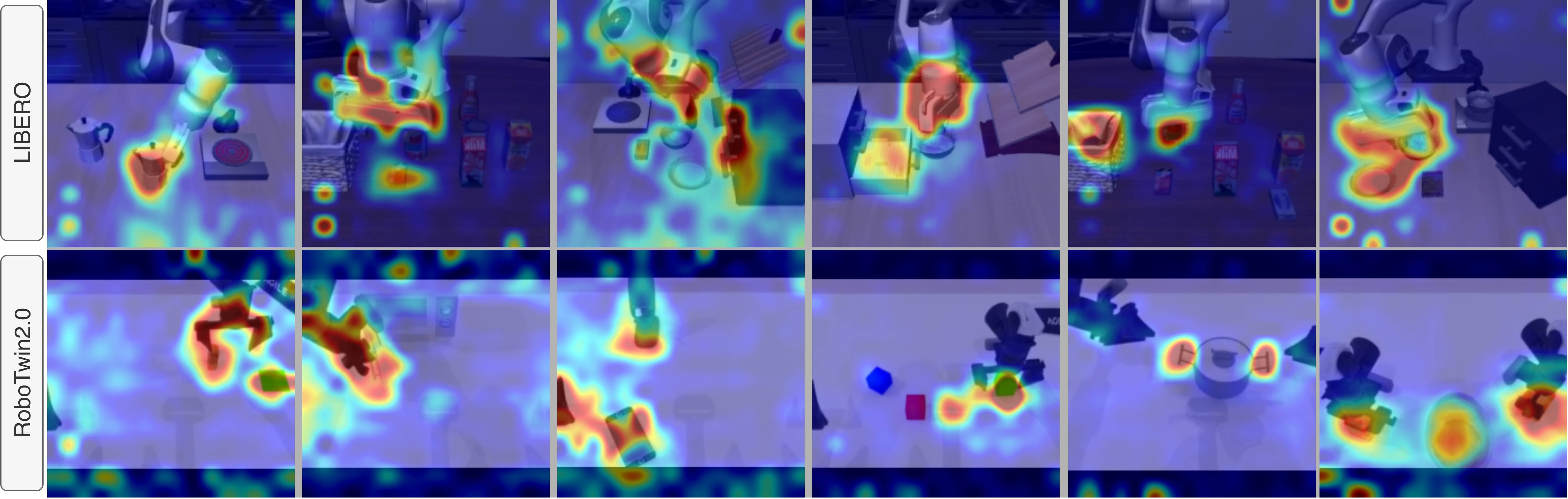}
\vspace{-6mm}
\caption{
\textbf{Visual attention of the Proposal Expert.}
Attention maps on LIBERO (top) and RoboTwin~2.0 (bottom); warmer colors highlight objects, interaction regions, and receptacles.
}
\label{fig:vis-attn}
\label{fig:vis-world1}
\end{minipage}\hfill%
\begin{minipage}[t]{0.318475\textwidth}
\vspace{0pt}
\centering
\includegraphics[width=\linewidth]{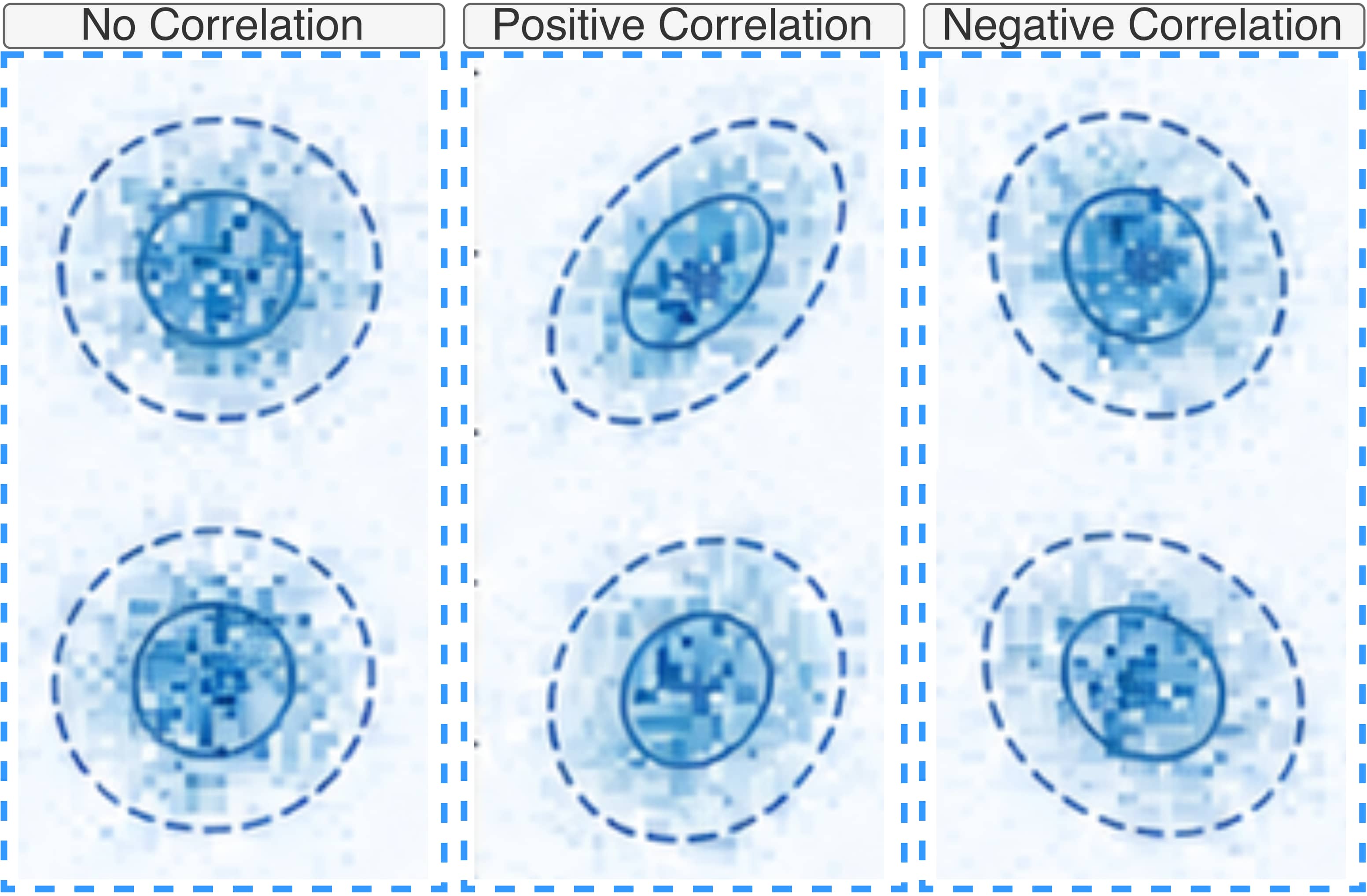}
\vspace{-6mm}
\caption{
\textbf{Calibrated source correlation.}
Action samples and correlation ellipses.
}
\label{fig:vis-world2}
\end{minipage}
\end{figure}

\begin{figure}[t]
\centering
\vspace{-4mm}
\includegraphics[width=1.0\textwidth]{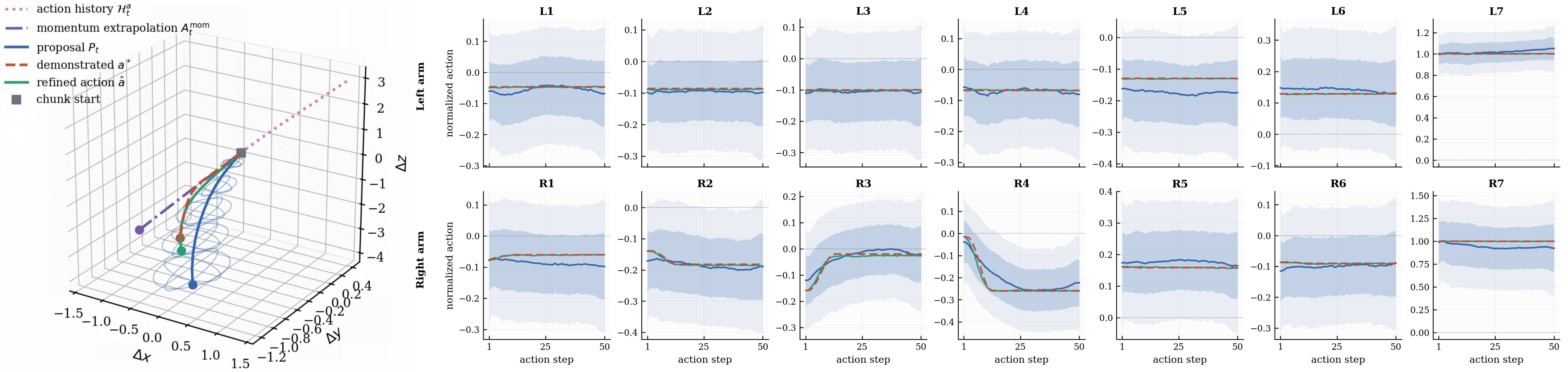}
\vspace{-6mm}
\caption{
\textbf{World-calibrated proposal refinement.}
Left: proposal refinement in 3D translation space. Right: normalized trajectories
and calibrated support across seven action dimensions.
}
\label{fig:vis-world}


\vspace{-3mm}
\end{figure}

\textbf{Scene-aware proposal grounding.}
Fig.~\ref{fig:vis-attn} shows that, despite compressed VLM KV caches, Proposal
Expert attention remains focused on manipulated objects and interaction regions
across LIBERO and RoboTwin~2.0 while suppressing background distractors. Thus,
cache compression preserves task cues for reliable scene-aware proposals during
execution.

\textbf{Calibrated source geometry.}
Fig.~\ref{fig:vis-world2} shows uncertainty over action dimensions. From left
to right, nearly circular contours indicate no correlation, upward-tilted
ellipses indicate positive correlation, and downward-tilted ellipses indicate
negative correlation. Dense samples mark source actions, while solid and dashed
ellipses denote approximately $1\sigma$ and $2\sigma$ contours. Together, the panels show how calibration adjusts the source distribution in extent, orientation, and cross-dimensional coupling according to action geometry.

\textbf{World-calibrated refinement.}
Fig.~\ref{fig:vis-world} shows that the proposal preserves recent motion's coarse
direction while refinement bends toward the demonstration within an anisotropic
region. Stable dimensions remain close to the proposal, whereas incompatible
ones move toward the desired future, consistent with the roles of extent and
geometry in Tab.~\ref{tab:source_ablation} across domains and robot embodiments.

\textbf{Efficient proposal-to-action transport.}
Fig.~\ref{fig:vis-world} and Tab.~\ref{tab:efficiency} show that calibrated
geometry shortens transport by starting near the demonstration and focusing
corrections along task-relevant directions, removing unnecessary transport
during generation. On LIBERO/real robots, latency falls by 25.8/16.3\%,
throughput rises by 34.8/19.5\%, and success improves by 1.5/8.7 points.

\vspace{-6pt}
\section{Related Work}
\label{sec:related_work}
\vspace{-3pt}

Beyond embodied control, multimodal learning has also explored fine-grained understanding and media manipulation grounding~\cite{peng2026finegrained, shao2024detecting, shao2023detecting}. These lines of work provide complementary perspectives on grounding visual semantics across modalities.

\textbf{Structured Priors for Action Generation.}
Recent generative robot policies increasingly inject structured priors to simplify action synthesis~\cite{hu2026abstraction, li2026global, li2026cogvla, li2026semanticvla}.
These priors improve task efficiency.
Diffusion and flow-based policies typically generate actions from a generic Gaussian source~\cite{chi2025diffusion,pi0,pi05,bjorck2025gr00t,zheng2026x,luo2026being}.
Recent work instead explores more informative starting points.
Streaming Flow Policy~\cite{jiang2025streaming} initializes generation around recently executed actions, while VITA~\cite{gao2026vita} directly transports visual latents toward action representations.
A2A~\cite{jia2026action} further encodes historical proprioceptive sequences as an informative flow source.
Together, these studies establish source design as an important dimension of generative action policies.
However, an informative source does not ensure that its implied motion remains suitable as the task evolves.
\method{} therefore treats recent motion as a proposal and determines how strongly and along which directions it should be corrected.

\textbf{World Models for Action Generation.}
A complementary line of work augments action policies with predictive world representations to account for future scene evolution~\cite{zhu2026delta, li2026unimpa, zhu2026h}.
DreamZero~\cite{ye2026world}, LingBot-VA~\cite{li2026causal}, DreamVLA~\cite{zhang2026dreamvla}, and WorldVLA~\cite{cen2025worldvla} couple future prediction or world modeling with action learning, while X-WAM~\cite{guo2026unified} incorporates future spatial structure through RGB-D prediction.
VLA-JEPA~\cite{sun2026vlajepa} introduces latent future prediction into VLA learning without explicit pixel-space generation, with related predictive objectives explored in UP-VLA~\cite{zhang2025up}.
Fast-WAM~\cite{yuan2026fast} further suggests that explicit test-time future generation is not always necessary.
Together, these studies leave open how predictive information should most effectively influence action generation.
In \method{}, the predicted future calibrates the proposal-centered source, controlling how far and along which directions the proposal is refined across diverse tasks, horizons, and embodiments.

\vspace{-6pt}
\section{Conclusion}
\label{sec:conclusion}
\vspace{-3pt}


We presented \method{}, a world-calibrated proposal-to-action framework that makes
the source of flow-based VLAs context dependent. A scene-aware motion proposal and
task-consistent future calibrate refinement extent and geometry while preserving the
original flow parameterization. Across simulation and real-world benchmarks, this
design improves robustness under perturbations and long-horizon execution while
reducing denoising and inference cost, motivating prospective source calibration
for robust and efficient generative robot policies across manipulation settings.

\section{AI Use Disclosure}
Generative AI tools were used solely to assist with English language polishing and to check formatting and presentation consistency in the manuscript. They were not used to develop research ideas, formulate hypotheses or mathematical claims, design the methodology or experiments, implement the proposed method, or analyze and interpret experimental results. All AI-assisted edits were reviewed by the authors, who take full responsibility for the final content of the paper.

\bibliography{iclr2027_conference}
\bibliographystyle{iclr2027_conference}
\newpage
\appendix
\noindent\textbf{The appendix is organized as follows:}
\begin{itemize}
    \item Sec.~\ref{app:implementation}: Implementation Details.
    \item Sec.~\ref{app:robotwin_full}: Supplementary RoboTwin~2.0 Results.
    \item Sec.~\ref{app:real_detail}: Real-World Evaluation Details.
    \item Sec.~\ref{app:quantitative_analysis}: Extended Qualitative Analysis.
    \item Sec.~\ref{app:derivations}: Derivations.
    \item Sec.~\ref{app:limitations_future}: Limitations and Future Work.
\end{itemize}

\section{Implementation Details}
\label{app:implementation}


\subsection{Model Details}
\label{app:model_details}

\subsubsection{Three-Layer KV Fusion: Instantiation}
\label{app:kv_fusion_impl}

With $L_v=18$ VLM layers and $L_p=6$ Proposal layers we set
$\rho_p(j)=3j-1$, so
$\mathcal{S}_j=\{3j-2,3j-1,3j\}$ splits the depth into six disjoint
centered windows of width $m^\star=1$; no boundary clamping is required.
Disjointness is a convenience, not a consequence of the analysis, which is
stated per window. See Appendix~\ref{app:kv_fusion} for the derivation of
the conditional risk characterization of $m^\star=1$ and the
inverse-variance weights $w_{t,j,l}$ used here.

\subsubsection{Calibrated Source: Efficient Evaluation and Sampling}
\label{app:calibration_impl}

\paragraph{Role of the action scale $D_a$.}
The released ProAct data pipeline applies the same coordinate-wise
normalization to the target action chunk and to the executed-action history.
For the ProAct configurations, this is quantile normalization: with empirical
quantiles $q_{.01,i}$ and $q_{.99,i}$ and
$\varepsilon_{\rm n}=10^{-6}$,
\begin{equation}
d_i=\frac{q_{.99,i}-q_{.01,i}+\varepsilon_{\rm n}}{2},
\qquad
\mu_{a,i}=q_{.01,i}+d_i,
\qquad
\bar a_i=\frac{a_i^{\rm raw}-\mu_{a,i}}{d_i}.
\label{eq:app_action_normalization}
\end{equation}
The supported z-score alternative has
$d_i=\operatorname{std}(a_i)+\varepsilon_{\rm n}$ and
$\mu_{a,i}=\operatorname{mean}(a_i)$. Hence $D_a$ is a fixed data-statistics
scale, not a learned model output. It has two roles. First, it converts the
raw residual to the dimensionless residual used by calibration:
$D_a^{-1}(a^{*,\rm raw}_{t,k}-P^{\rm raw}_{t,k})
=\bar a^*_{t,k}-\bar P_{t,k}$. The additive center cancels, so it never enters
the extent or geometry losses. Second, it maps a covariance learned in
normalized coordinates back to original units by congruence,
$C^a_{t,k}=D_aC_{t,k}D_a^\top$.

Importantly, normalization is performed before the ProAct model is called and
unnormalization after it returns. Thus the implementation stores
$\bar a^*$, $\bar P$, and $\bar X_{\rm src}$ and does not explicitly construct
or apply $D_a$ inside the model; internally, its role is already absorbed and
is equivalent to setting $D_a=I$. In particular, the code expression
\texttt{actions - proposal} is exactly the normalized residual above, not an
unnormalized residual missing a factor of $D_a^{-1}$. Likewise, multiplying
the sampled source by $D_a$ inside the model would incorrectly apply the action
scale twice.

The geometry objective of \eqref{eq:geometry_loss} uses the normalized
shape covariance $\widetilde C_{t,k}$. The model samples the source using
$C_{t,k}$ in normalized coordinates; $C_{t,k}^a=D_aC_{t,k}D_a^\top$ is its
original-coordinate interpretation after output unnormalization. Neither
requires forming a dense $d_a\times d_a$ matrix. Let
$\alpha_{t,k}=d_a/\operatorname{tr}(I+B_{t,k}B_{t,k}^{\top})$, so that
$\widetilde C_{t,k}=\alpha_{t,k}(I+B_{t,k}B_{t,k}^{\top})$.
The Woodbury identity and the matrix determinant lemma give
\begin{equation}
\widetilde C_{t,k}^{-1}
=
\alpha_{t,k}^{-1}\!\left(
I-B_{t,k}(I_r+B_{t,k}^{\top}B_{t,k})^{-1}B_{t,k}^{\top}
\right),
\end{equation}
\begin{equation}
\log\det\widetilde C_{t,k}
=
d_a\log\alpha_{t,k}
+
\log\det(I_r+B_{t,k}^{\top}B_{t,k}),
\label{eq:app_woodbury}
\end{equation}
so both terms cost $O(d_ar^2+r^3)=O(d_ar^2)$ for $r\leq d_a$ and never
require a $d_a\times d_a$ inverse.

Sampling is equally cheap: for each $k$, draw mutually independent
$\epsilon_{a,k}\sim\mathcal{N}(0,I_{d_a})$ and
$\xi_{t,k}\sim\mathcal{N}(0,I_r)$,
\begin{equation}
\bar X_{t,\mathrm{src},k}
=
\bar P_{t,k}
+
\sigma_{t,k}\sqrt{\alpha_{t,k}}\,
\left(\epsilon_{a,k}+B_{t,k}\xi_{t,k}\right),
\label{eq:app_sampling}
\end{equation}
which realizes $C_{t,k}$ without any Cholesky factorization. Applying the
output transform gives
\begin{equation}
X_{t,\mathrm{src},k}^{a}
=\mu_a+D_a\bar X_{t,\mathrm{src},k}
=P_{t,k}^{\rm raw}
+\sigma_{t,k}\sqrt{\alpha_{t,k}}\,
D_a\!\left(\epsilon_{a,k}+B_{t,k}\xi_{t,k}\right),
\end{equation}
and therefore
$\operatorname{Cov}(X_{t,\mathrm{src},k}^{a}\mid P_t,U_t,B_t)
=C_{t,k}^a=D_aC_{t,k}D_a^\top$.
The block-diagonal Gaussian construction makes the $H$ action steps
conditionally independent given $(P_t,U_t,B_t)$. Padding dimensions of the
action tensor are excluded from
$e_{t,k}$, $\mathcal{L}_U$, and $\mathcal{L}_B$, and $d_a$ counts only the
effective action dimensions. See Appendix~\ref{app:calibration} for why the
projection this evaluation relies on ($S\le S_{\max}$) is necessary.

\subsubsection{Action Expert Architecture and Inference}
\label{app:action_expert}

The Action Expert keeps the layer count, width, and adaRMS time conditioning
of the base VLA action expert. At layer $l$, its $H$ action-state tokens attend
jointly to three sources of keys and values,
\begin{equation}
\overline H_{t,s,l}^{a}
=
\Attn\!\left(
Q_l^{a}(H_{t,s,l-1}^{a}),
\left[K_{t,l}^{v};K_{l}^{w};K_{t,s,l}^{a}\right],
\left[V_{t,l}^{v};V_{l}^{w};V_{t,s,l}^{a}\right]
\right),
\label{eq:app_action_attention}
\end{equation}
where $K_{t,l}^{v},V_{t,l}^{v}$ is the VLM cache at the aligned depth,
$K_{l}^{w},V_{l}^{w}$ are layer-wise projections of the
$H\times N_z$ world tokens $H_t^w$, and
$K_{t,s,l}^{a},V_{t,s,l}^{a}$ are its own projections.

\textbf{Training-time convention.}
In \eqref{eq:action_flow}, a demonstration index and valid time index are
sampled from the empirical training distribution,
$s\sim\operatorname{Uniform}[0,1]$, and $\epsilon_p,\epsilon_a$ are independent
standard Gaussian noises independent of $s$. In the released configuration,
the proposal, extent, and geometry are stop-gradient quantities when forming
$X_{t,\mathrm{src}}^a$; consequently, both this source and
$v_t^*=X_{t,\mathrm{src}}^a-a_t^*$ are held fixed with respect to their
source-producing branches under $\mathcal{L}_{\rm flow}$. The flow loss still
trains the Action Expert and can train the World Expert through the separate
world-context conditioning path. The proposal, future, extent, and geometry
objectives provide the corresponding direct supervision to the preceding
modules. Thus \eqref{eq:action_flow} is standard conditional flow
regression with a fixed sampled source, rather than an objective that can be
reduced by moving the source itself.
For completeness, the extent loss is
$\mathcal{L}_U=H^{-1}\sum_{k=1}^{H}(U_{t,k}-U_{t,k}^*)^2$.

Because the VLM prefix never attends to the action stream, the VLM cache and
the world tokens are independent of the flow time $s$; both are computed once
per observation and reused across all integration steps. Hence, the added
inference cost over the base policy is one Proposal Expert pass and one World
Expert pass. Integration follows the base policy exactly: starting from
$X_{t,\mathrm{src}}^{a}$ at $s=1$, we take uniform Euler steps
\begin{equation}
X^{a}
\leftarrow
X^{a}
-
\frac{1}{N_{\mathrm{step}}}v_\theta^A
\end{equation}
down to $s=0$.


\newcolumntype{Y}{>{\centering\arraybackslash}X}

\begin{table*}[t]
\vspace{-6mm}
\centering
\caption{\textbf{Training Configuration of \method{}.}}
\label{tab:training_configuration}

\footnotesize
\setlength{\tabcolsep}{5.0pt}
\renewcommand{\arraystretch}{1.10}

\begin{tabularx}{0.75\textwidth}{@{}lYYY@{}}
\toprule

\textbf{Setting}
& \textbf{LIBERO}
& \textbf{RoboTwin 2.0}
& \textbf{Real-World} \\

\midrule
\multicolumn{4}{@{}l}{\textit{\textbf{Optimization}}} \\
\midrule

Training steps
& 60k
& 30k
& 30k \\

Batch size
& 128
& 64
& 64 \\

Optimizer
& \multicolumn{3}{c}{AdamW, $\beta_1=0.9$, $\beta_2=0.95$} \\

Peak learning rate
& $5\times10^{-5}$
& $5\times10^{-5}$
& $5\times10^{-5}$ \\

Warm-up steps
& 10k
& 10k
& 10k \\

GPUs
& 4
& 4
& 4 \\

\midrule
\multicolumn{4}{@{}l}{
\textit{\textbf{Loss Weights (Eq.~\ref{eq:total_loss})}}
} \\
\midrule

$\lambda_p$ (proposal)
& 0.1
& 0.1
& 0.1 \\

$\lambda_f$ (future)
& 0.01
& 0.01
& 0.01 \\

$\lambda_u$ (extent)
& 0.1
& 0.1
& 0.1 \\

$\lambda_b$ (geometry)
& 0.001
& 0.001
& 0.001 \\

$\lambda_a$ (flow)
& 1.0
& 1.0
& 1.0 \\

\midrule
\multicolumn{4}{@{}l}{
\textit{\textbf{Architecture and Calibration}}
} \\
\midrule

Action horizon $H$
& 10
& 50
& 50 \\

Action history length $M$
& 10
& 50
& 50 \\

Proposal Expert depth $L_p$
& \multicolumn{3}{c}{6} \\

World queries $N_z$
& 16
& 4
& 4 \\

Geometry rank $r$
& \multicolumn{3}{c}{4} \\

Geometry budget $S_{\max}$
& \multicolumn{3}{c}{5.0} \\

$\tau_u$
& \multicolumn{3}{c}{0.35} \\

$\sigma_{\min}$, $\sigma_{\max}$
& \multicolumn{3}{c}{0.05, 1.0} \\

Momentum $\beta$, $\lambda$
& \multicolumn{3}{c}{0.9, 0.9} \\

\midrule
\multicolumn{4}{@{}l}{
\textit{\textbf{Data}}
} \\
\midrule

Input cameras
& 2
& 3
& 3 \\

Input resolution
& \multicolumn{3}{c}{$224\times224$} \\

Number of arms
& 1
& 2
& 2 \\

Robotic arm type
& Franka Emika Panda
& Aloha-AgileX
& Galaxea R1 Lite \& Aloha-AgileX \\

\bottomrule
\end{tabularx}

\end{table*}

\subsection{Training Details}
\label{app:training_details}

\subsubsection{Episode Boundary Handling}
\label{app:episode_boundaries}

Two boundary cases are handled by masking rather than by discarding samples.
Near the end of an episode, the future clip is completed by repeating the last
available frame. Because temporal attention would allow these padded frames to
contaminate every latent token, such samples are excluded from
$\mathcal{L}_{\mathrm{future}}$ while keeping all remaining losses.

Near the beginning of an episode, the action history is shorter than $M$.
The missing entries and the motion state are initialized to zero, while
gripper dimensions are initialized to the gripper value of the first executed
action. At deployment, the motion state is recomputed from scratch over the
last $M$ executed actions at every call, ensuring that the analytical proposal
is identical during training and inference for consistency across repeated evaluations.

\subsubsection{Training Configuration}
\label{app:train_config}
Tab.~\ref{tab:training_configuration} summarizes the optimization, loss,
architecture, and data configurations used for \method{}.
We use the same optimizer and peak learning rate across LIBERO, RoboTwin~2.0,
and real-world experiments, while adapting the training schedule and batch size
to the scale of each setting. The action history length is matched to the
action horizon ($M=H$), with $H=10$ for LIBERO and $H=50$ for both RoboTwin~2.0
and real-world experiments. Most architectural and calibration hyperparameters,
including the Proposal Expert depth, geometry rank, geometry budget, extent
threshold, and momentum coefficients, are shared across all settings; only the
number of world queries is adjusted, with $N_z=16$ for LIBERO and $N_z=4$ for
RoboTwin~2.0 and real-world experiments. The real-world configuration otherwise
follows the RoboTwin~2.0 setting.

\section{Supplementary RoboTwin~2.0 Results}

\newlength{\RTMethodW}
\newlength{\RTTaskW}
\newlength{\RTTaskWLast}

\setlength{\RTMethodW}{3.20cm}

\begin{table*}[p]
\centering
\vspace*{-0.75cm}
\caption{
\textbf{Per-task success rates on 50 RoboTwin~2.0 tasks under
Randomized (Hard).}
All methods use the same protocol.
The overall average is reported in the final block.
}
\label{tab:robotwin_random_50results}

\tiny
\setlength{\tabcolsep}{1.5pt}
\renewcommand{\arraystretch}{0.92}

\setlength{\RTTaskW}{%
  \dimexpr
  (\textwidth-\RTMethodW-20\tabcolsep-\arrayrulewidth)/10
  \relax
}

\setlength{\RTTaskWLast}{%
  \dimexpr
  (\textwidth-\RTMethodW-22\tabcolsep-\arrayrulewidth)/11
  \relax
}


\begin{tabular}{@{}>{\raggedright\arraybackslash}m{\RTMethodW}|*{10}{>{\centering\arraybackslash}p{\RTTaskW}}@{}}
\toprule

\raisebox{1.15ex}{\textbf{Method}}
& \shortstack{Adjust\\[-1pt]Bottle}
& \shortstack{Beat\\[-1pt]Block\\[-1pt]Hammer}
& \shortstack{Blocks\\[-1pt]Rank.\\[-1pt]RGB}
& \shortstack{Blocks\\[-1pt]Rank.\\[-1pt]Size}
& \shortstack{Click\\[-1pt]Alarmclock}
& \shortstack{Click\\[-1pt]Bell}
& \shortstack{Dump\\[-1pt]Bin\\[-1pt]Bigbin}
& \shortstack{Grab\\[-1pt]Roller}
& \shortstack{Handover\\[-1pt]Block}
& \shortstack{Handover\\[-1pt]Mic} \\
\midrule

DP3~\cite{ze20243d}
& 3\% & 8\% & 0\% & 0\% & 14\% & 0\% & 53\% & 2\% & 0\% & 3\% \\

$\pi_0$~\cite{pi0}
& 56\% & 21\% & 5\% & 1\% & 11\% & 3\% & 24\% & 80\% & 8\% & 13\% \\

$\pi_{0.5}$~\cite{pi05}
& \textbf{83\%}
& \textbf{50\%}
& 35\%
& 14\%
& \textbf{93\%}
& \textbf{76\%}
& 42\%
& 89\%
& 19\%
& 18\% \\

X-VLA~\cite{zheng2026x}
& 33\% & 9\% & 4\% & \textbf{25\%} & 43\% & 63\% & 41\% & 20\% & 1.3\% & 1\% \\

Spatial Forcing~\cite{li2026spatial}
& 71\% & 13\% & 7\% & 3\% & 28\% & 28\% & 70\% & \textbf{94\%} & 2\% & 2\% \\

X-WAM~\cite{guo2026unified}
& 13\% & 21\% & 36\% & 0\% & 61\% & 52\% & 40\% & 16\% & \textbf{21\%} & 8\% \\

Abot-M0~\cite{yang2026abot}
& 49\% & 33\% & 23\% & 7\% & 21\% & 11\% & 68\% & 91\% & 0\% & \textbf{22\%} \\

EventVLA~\cite{yang2026eventvla}
& 24\% & 1\% & 6\% & 1\% & 30\% & 48\% & 64\% & 48\% & 4\% & 5\% \\

Xiaomi Robotics-0~\cite{cai2026xiaomi}
& 9\% & 16\% & 8\% & 1\% & 58\% & 57\% & 49\% & 12\% & 1\% & 6\% \\

FastWAM~\cite{yuan2026fast}
& 0\% & 0\% & 0\% & 0\% & 60\% & 7\% & 0\% & 2\% & 0\% & 0\% \\

GalaxeaVLA~\cite{liu2026g05autoregressivestreamrobot}
& 10\% & 31\% & 1\% & 0\% & 7\% & 27\% & 40\% & 33\% & 0\% & 2\% \\

AHA-WAM~\cite{cai2026aha}
& 0\% & 0\% & 1\% & 0\% & 17\% & 19\% & 27\% & 3\% & 1\% & 3\% \\

starVLA~\cite{community2026starvla}
& 0\% & 2\% & 2\% & 0\% & 0\% & 7\% & 4\% & 44\% & 0\% & 1\% \\

\midrule
\rowcolor[HTML]{DDE7F4}
\textbf{\method{}}
& 75\%
& 28\%
& \textbf{46\%}
& 20\%
& 61\%
& 50\%
& \textbf{78\%}
& \textbf{94\%}
& 11\%
& 10\% \\

\bottomrule
\end{tabular}

\vspace{0.2em}


\begin{tabular}{@{}>{\raggedright\arraybackslash}m{\RTMethodW}|*{10}{>{\centering\arraybackslash}p{\RTTaskW}}@{}}
\toprule

\raisebox{1.15ex}{\textbf{Method}}
& \shortstack{Hanging\\[-1pt]Mug}
& \shortstack{Lift\\[-1pt]Pot}
& \shortstack{Move\\[-1pt]Can\\[-1pt]Pot}
& \shortstack{Move\\[-1pt]Pillbottle\\[-1pt]Pad}
& \shortstack{Move\\[-1pt]Playingcard\\[-1pt]Away}
& \shortstack{Move\\[-1pt]Stapler\\[-1pt]Pad}
& \shortstack{Open\\[-1pt]Laptop}
& \shortstack{Open\\[-1pt]Microwave}
& \shortstack{Pick\\[-1pt]Diverse\\[-1pt]Bottles}
& \shortstack{Pick\\[-1pt]Dual\\[-1pt]Bottles} \\
\midrule

DP3~\cite{ze20243d}
& 1\% & 0\% & 6\% & 0\% & 3\% & 0\% & 7\% & 22\% & 1\% & 1\% \\

$\pi_0$~\cite{pi0}
& 3\% & \textbf{36\%} & 21\% & 1\% & 22\% & 2\% & 46\% & 50\% & 6\% & 12\% \\

$\pi_{0.5}$~\cite{pi05}
& 3\%
& 0\%
& 27\%
& \textbf{29\%}
& \textbf{67\%}
& \textbf{18\%}
& 35\%
& 37\%
& 3\%
& 6\% \\

X-VLA~\cite{zheng2026x}
& 2\% & 15\% & 0\% & 3\% & 5\% & 0\% & 16\% & 5\% & 30\% & 51\% \\

Spatial Forcing~\cite{li2026spatial}
& 1\% & 0\% & 1\% & 14\% & 15\% & 1\% & 15\% & 18\% & 38\% & 47\% \\

X-WAM~\cite{guo2026unified}
& 0\% & 0\% & \textbf{34\%} & 26\% & 11\% & 2\% & 6\% & 24\% & 26\% & 55\% \\

Abot-M0~\cite{yang2026abot}
& 4\% & 29\% & 4\% & 4\% & 50\% & 4\% & 64\% & 26\% & 7\% & 19\% \\

EventVLA~\cite{yang2026eventvla}
& 2\% & 1\% & 0\% & 3\% & 17\% & 2\% & 23\% & 44\% & 9\% & 8\% \\

Xiaomi Robotics-0~\cite{cai2026xiaomi}
& 6\% & 1\% & 7\% & 3\% & 12\% & 0\% & 24\% & 16\% & 12\% & 28\% \\

FastWAM~\cite{yuan2026fast}
& 0\% & 0\% & 0\% & 0\% & 0\% & 0\% & 0\% & 0\% & 0\% & 1\% \\

GalaxeaVLA~\cite{liu2026g05autoregressivestreamrobot}
& 1\% & 0\% & 5\% & 2\% & 13\% & 1\% & 13\% & 13\% & 11\% & 21\% \\

AHA-WAM~\cite{cai2026aha}
& 0\% & 0\% & 0\% & 0\% & 0\% & 0\% & 9\% & 1\% & 0\% & 1\% \\

starVLA~\cite{community2026starvla}
& 0\% & 0\% & 0\% & 0\% & 0\% & 0\% & 2\% & 1\% & 15\% & 20\% \\

\midrule
\rowcolor[HTML]{DDE7F4}
\textbf{\method{}}
& \textbf{10\%}
& 33\%
& 27\%
& \textbf{29\%}
& 55\%
& 13\%
& \textbf{76\%}
& \textbf{54\%}
& \textbf{57\%}
& \textbf{63\%} \\

\bottomrule
\end{tabular}

\vspace{0.2em}


\begin{tabular}{@{}>{\raggedright\arraybackslash}m{\RTMethodW}|*{10}{>{\centering\arraybackslash}p{\RTTaskW}}@{}}
\toprule

\raisebox{1.15ex}{\textbf{Method}}
& \shortstack{Place\\[-1pt]A2B\\[-1pt]Left}
& \shortstack{Place\\[-1pt]A2B\\[-1pt]Right}
& \shortstack{Place\\[-1pt]Bread\\[-1pt]Basket}
& \shortstack{Place\\[-1pt]Bread\\[-1pt]Skillet}
& \shortstack{Place\\[-1pt]Burger\\[-1pt]Fries}
& \shortstack{Place\\[-1pt]Can\\[-1pt]Basket}
& \shortstack{Place\\[-1pt]Cans\\[-1pt]Plasticbox}
& \shortstack{Place\\[-1pt]Container\\[-1pt]Plate}
& \shortstack{Place\\[-1pt]Dual\\[-1pt]Shoes}
& \shortstack{Place\\[-1pt]Empty\\[-1pt]Cup} \\
\midrule

DP3~\cite{ze20243d}
& 2\% & 0\% & 1\% & 0\% & 18\% & 2\% & 3\% & 1\% & 0\% & 1\% \\

$\pi_0$~\cite{pi0}
& 1\% & 6\% & 4\% & 1\% & 4\% & 5\% & 2\% & 45\% & 0\% & 11\% \\

$\pi_{0.5}$~\cite{pi05}
& \textbf{60\%}
& \textbf{57\%}
& 56\%
& 46\%
& 70\%
& \textbf{25\%}
& 47\%
& \textbf{78\%}
& 7\%
& \textbf{86\%} \\

X-VLA~\cite{zheng2026x}
& 15\% & 13\% & 52\% & 17\% & 35\% & 9\% & 60\% & 39\% & 12\% & 16\% \\

Spatial Forcing~\cite{li2026spatial}
& 11\% & 6\% & 53\% & 30\% & 71\% & 10\% & 11\% & 54\% & 2\% & 33\% \\

X-WAM~\cite{guo2026unified}
& 18\% & 21\% & 40\% & 37\% & 36\% & 15\% & 36\% & 52\% & 7\% & 36\% \\

Abot-M0~\cite{yang2026abot}
& 26\% & 22\% & 47\% & 25\% & 55\% & 13\% & 36\% & 62\% & 15\% & 43\% \\

EventVLA~\cite{yang2026eventvla}
& 20\% & 17\% & 15\% & 4\% & 7\% & 7\% & 14\% & 15\% & 1\% & 18\% \\

Xiaomi Robotics-0~\cite{cai2026xiaomi}
& 6\% & 2\% & 23\% & 11\% & 27\% & 7\% & 31\% & 59\% & 9\% & 12\% \\

FastWAM~\cite{yuan2026fast}
& 0\% & 1\% & 0\% & 0\% & 0\% & 0\% & 0\% & 0\% & 0\% & 0\% \\

GalaxeaVLA~\cite{liu2026g05autoregressivestreamrobot}
& 5\% & 3\% & 32\% & 14\% & 32\% & 6\% & 22\% & 38\% & 9\% & 3\% \\

AHA-WAM~\cite{cai2026aha}
& 1\% & 0\% & 3\% & 0\% & 0\% & 2\% & 1\% & 11\% & 0\% & 2\% \\

starVLA~\cite{community2026starvla}
& 0\% & 1\% & 5\% & 2\% & 11\% & 0\% & 2\% & 3\% & 2\% & 0\% \\

\midrule
\rowcolor[HTML]{DDE7F4}
\textbf{\method{}}
& 50\%
& 44\%
& \textbf{62\%}
& \textbf{48\%}
& \textbf{84\%}
& 11\%
& \textbf{63\%}
& 76\%
& \textbf{39\%}
& 82\% \\

\bottomrule
\end{tabular}

\vspace{0.2em}


\begin{tabular}{@{}>{\raggedright\arraybackslash}m{\RTMethodW}|*{10}{>{\centering\arraybackslash}p{\RTTaskW}}@{}}
\toprule

\raisebox{1.15ex}{\textbf{Method}}
& \shortstack{Place\\[-1pt]Fan}
& \shortstack{Place\\[-1pt]Mouse\\[-1pt]Pad}
& \shortstack{Place\\[-1pt]Object\\[-1pt]Basket}
& \shortstack{Place\\[-1pt]Object\\[-1pt]Scale}
& \shortstack{Place\\[-1pt]Object\\[-1pt]Stand}
& \shortstack{Place\\[-1pt]Phone\\[-1pt]Stand}
& \shortstack{Place\\[-1pt]Shoe}
& \shortstack{Press\\[-1pt]Stapler}
& \shortstack{Put\\[-1pt]Bottles\\[-1pt]Dustbin}
& \shortstack{Put\\[-1pt]Object\\[-1pt]Cabinet} \\
\midrule

DP3~\cite{ze20243d}
& 1\% & 1\% & 0\% & 0\% & 0\% & 2\% & 2\% & 3\% & 21\% & 1\% \\

$\pi_0$~\cite{pi0}
& 10\% & 1\% & 2\% & 0\% & 11\% & 7\% & 6\% & 29\% & 13\% & 18\% \\

$\pi_{0.5}$~\cite{pi05}
& \textbf{36\%}
& \textbf{26\%}
& \textbf{36\%}
& \textbf{49\%}
& \textbf{65\%}
& \textbf{53\%}
& \textbf{61\%}
& 70\%
& 9\%
& 15\% \\

X-VLA~\cite{zheng2026x}
& 1\% & 0\% & 6\% & 6\% & 43\% & 5\% & 39\% & \textbf{73\%} & 20\% & 21\% \\

Spatial Forcing~\cite{li2026spatial}
& 7\% & 9\% & 16\% & 4\% & 32\% & 1\% & \textbf{61\%} & 55\% & 41\% & 1\% \\

X-WAM~\cite{guo2026unified}
& 6\% & 6\% & 20\% & 14\% & 28\% & 4\% & 31\% & 20\% & 21\% & 16\% \\

Abot-M0~\cite{yang2026abot}
& 10\% & 5\% & 33\% & 23\% & 25\% & 26\% & 24\% & 52\% & 26\% & 22\% \\

EventVLA~\cite{yang2026eventvla}
& 5\% & 6\% & 29\% & 24\% & 32\% & 3\% & 10\% & 32\% & 31\% & \textbf{27\%} \\

Xiaomi Robotics-0~\cite{cai2026xiaomi}
& 3\% & 1\% & 15\% & 2\% & 24\% & 11\% & 29\% & 21\% & 21\% & 5\% \\

FastWAM~\cite{yuan2026fast}
& 0\% & 0\% & 0\% & 0\% & 0\% & 0\% & 0\% & 3\% & 0\% & 0\% \\

GalaxeaVLA~\cite{liu2026g05autoregressivestreamrobot}
& 2\% & 1\% & 3\% & 2\% & 27\% & 8\% & 22\% & 17\% & 3\% & 5\% \\

AHA-WAM~\cite{cai2026aha}
& 0\% & 0\% & 0\% & 0\% & 0\% & 0\% & 5\% & 2\% & 1\% & 0\% \\

starVLA~\cite{community2026starvla}
& 1\% & 0\% & 1\% & 1\% & 0\% & 0\% & 0\% & 10\% & 0\% & 0\% \\

\midrule
\rowcolor[HTML]{DDE7F4}
\textbf{\method{}}
& 28\%
& 25\%
& 35\%
& 34\%
& 61\%
& 32\%
& 53\%
& 57\%
& \textbf{69\%}
& 21\% \\

\bottomrule
\end{tabular}

\vspace{0.2em}


\begin{tabular}{@{}>{\raggedright\arraybackslash}m{\RTMethodW}|*{11}{>{\centering\arraybackslash}p{\RTTaskWLast}}@{}}
\toprule

\raisebox{1.15ex}{\textbf{Method}}
& \shortstack{Rotate\\[-1pt]QRCode}
& \shortstack{Scan\\[-1pt]Object}
& \shortstack{Shake\\[-1pt]Bottle\\[-1pt]Horiz.}
& \shortstack{Shake\\[-1pt]Bottle}
& \shortstack{Stack\\[-1pt]Blocks\\[-1pt]Three}
& \shortstack{Stack\\[-1pt]Blocks\\[-1pt]Two}
& \shortstack{Stack\\[-1pt]Bowls\\[-1pt]Three}
& \shortstack{Stack\\[-1pt]Bowls\\[-1pt]Two}
& \shortstack{Stamp\\[-1pt]Seal}
& \shortstack{Turn\\[-1pt]Switch}
& \shortstack{\textbf{Overall}\\[-1pt]\textbf{Avg.}} \\
\midrule

DP3~\cite{ze20243d}
& 1\% & 1\% & 25\% & 19\% & 0\% & 0\% & 5\% & 6\% & 0\% & 8\% & 4.96\% \\

$\pi_0$~\cite{pi0}
& 15\% & 1\% & 51\% & 60\% & 0\% & 1\% & 24\% & 41\% & 4\% & \textbf{23\%} & 16.34\% \\

$\pi_{0.5}$~\cite{pi05}
& \textbf{56\%}
& \textbf{38\%}
& \textbf{100\%}
& \textbf{100\%}
& 16\%
& 56\%
& 35\%
& 66\%
& 23\%
& 6\%
& 43.84\% \\

X-VLA~\cite{zheng2026x}
& 1\% & 9\% & 78\% & 76\% & 0\% & 2\% & 9\% & 15\% & 2\% & 6\% & 20.9\% \\

Spatial Forcing~\cite{li2026spatial}
& 22\% & 6\% & 87\% & 84\% & 0\% & 19\% & \textbf{55\%} & \textbf{79\%} & 2\% & 7\% & 26.7\% \\

X-WAM~\cite{guo2026unified}
& 2\% & 14\% & 66\% & 67\% & \textbf{26\%} & \textbf{76\%} & 38\% & 67\% & 16\% & 4\% & 25.8\% \\

Abot-M0~\cite{yang2026abot}
& 6\% & 17\% & 85\% & 84\% & 10\% & 42\% & 44\% & 65\% & 19\% & 20\% & 30.4\% \\

EventVLA~\cite{yang2026eventvla}
& 1\% & 3\% & 37\% & 34\% & 3\% & 13\% & 11\% & 17\% & 3\% & 8\% & 15.7\% \\

Xiaomi Robotics-0~\cite{cai2026xiaomi}
& 1\% & 8\% & 62\% & 61\% & 7\% & 22\% & 35\% & 53\% & 3\% & 14\% & 18.2\% \\

FastWAM~\cite{yuan2026fast}
& 0\% & 0\% & 9\% & 10\% & 0\% & 0\% & 0\% & 0\% & 0\% & 3\% & 1.9\% \\

GalaxeaVLA~\cite{liu2026g05autoregressivestreamrobot}
& 2\% & 6\% & 46\% & 45\% & 1\% & 9\% & 5\% & 27\% & 2\% & 8\% & 12.7\% \\

AHA-WAM~\cite{cai2026aha}
& 0\% & 0\% & 15\% & 14\% & 1\% & 2\% & 0\% & 5\% & 0\% & 13\% & 3.2\% \\

starVLA~\cite{community2026starvla}
& 0\% & 5\% & 9\% & 5\% & 1\% & 1\% & 0\% & 0\% & 0\% & 0\% & 3.2\% \\

\midrule
\rowcolor[HTML]{DDE7F4}
\textbf{\method{}}
& 15\%
& 34\%
& \textbf{100\%}
& \textbf{100\%}
& \textbf{26\%}
& 44\%
& 48\%
& 73\%
& \textbf{29\%}
& 20\%
& \textbf{47.06\%} \\

\bottomrule
\end{tabular}

\end{table*}
\label{app:robotwin_full}

To complement the representative results in the main paper, we provide a
task-level evaluation over all 50 RoboTwin~2.0 tasks under the Randomized
(Hard) setting in Tab.~\ref{tab:robotwin_random_50results}. Each task is
evaluated with 100 rollouts under the same protocol. The full benchmark covers
a diverse range of manipulation behaviors, including interaction, placement,
handover, articulated-object manipulation, and multi-object coordination,
thereby providing a more comprehensive assessment of robustness under
substantial visual and physical variation.

Across all 50 tasks, \method{} achieves the highest average success rate of
\textbf{47.06\%}, exceeding the strong $\pi_{0.5}$ baseline (43.84\%) by
\textbf{3.22 percentage points}. It also substantially outperforms Abot-M0
(30.4\%), Spatial Forcing (26.7\%), X-WAM (25.8\%), and X-VLA (20.9\%).
At the task level, \method{} improves over $\pi_{0.5}$ on 22 tasks and ties
on another four, while achieving the best or tied-best performance among all
compared methods on 19 tasks. Notably, the improvements are particularly large
on several challenging tasks, including \textit{Put Bottles Dustbin}
(69\% vs.\ 9\%), \textit{Pick Dual Bottles} (63\% vs.\ 6\%),
\textit{Pick Diverse Bottles} (57\% vs.\ 3\%), \textit{Open Laptop}
(76\% vs.\ 35\%), and \textit{Dump Bin Bigbin} (78\% vs.\ 42\%).
\method{} also retains strong performance on high-success tasks, reaching
100\% on both bottle-shaking variants and 94\% on \textit{Grab Roller}.
Although the relative advantage varies across individual tasks, the aggregate
results show that the proposal--calibrate--refine formulation provides a
stronger overall policy across the heterogeneous RoboTwin~2.0 task
distribution.

\begin{figure}[t]
  \centering
  \includegraphics[width=\textwidth]{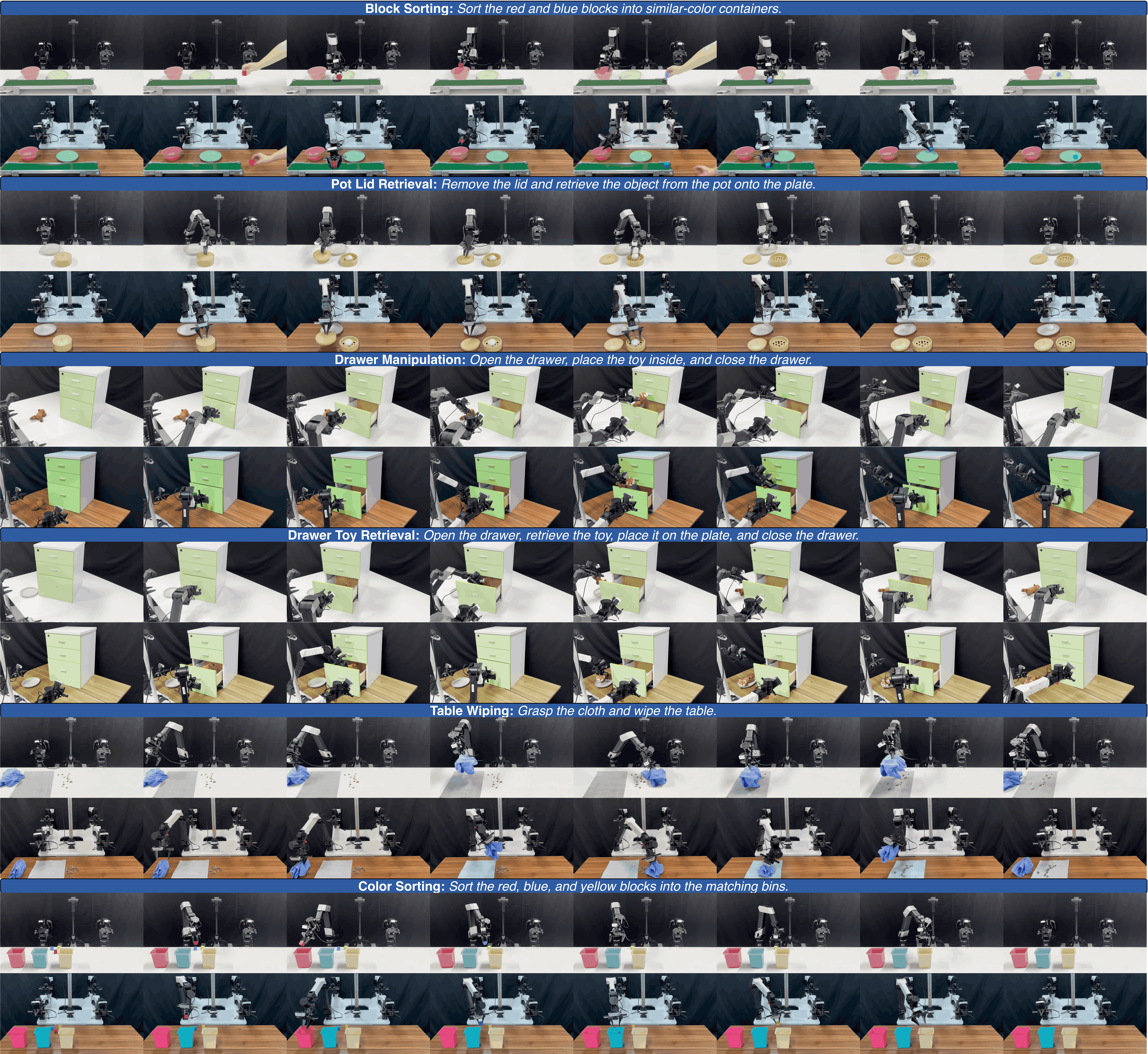}
  \caption{Representative successful real-world rollouts of \method{} on the
  six evaluation tasks. Within each task panel, the top row shows GALAXEA R1
  Lite and the bottom row shows AgileX Cobot Magic; snapshots in each row are
  ordered from left to right. The eight snapshots are selected at semantic
  milestones rather than at uniform temporal intervals, covering repeated
  sorting, container and drawer manipulation, object transfer, and sustained
  wiping.}
  \label{fig:real-rollouts}
\end{figure}

\section{Real-World Evaluation Details}
\label{app:real_detail}
\subsection{Evaluation Protocol and Metrics}
\label{app:real_metrics}

We use a unified evaluation setup for all methods. Each policy is tested with the same task instructions, sensing configuration, camera placement, control interface, and randomized initialization protocol. Unless stated otherwise, every task contains $N=25$ independent trials. Object and robot initial states are sampled from task-specific admissible ranges that are fixed before evaluation and shared across methods. No human correction is allowed once a trial starts. 

Each task is represented by an ordered sequence of $S_i$ checkpoints. Let $c_{i,j}^{(n)}\in\{0,1\}$ denote whether trial $n$ reaches checkpoint $j$ of task $i$. To enforce sequential completion, we evaluate checkpoint $j$ through the progressive indicator
\begin{equation}
\small
\bar{c}_{i,j}^{(n)}
=
\prod_{k=1}^{j} c_{i,k}^{(n)},
\qquad
\mathrm{SR}_{i,j}
=
\frac{1}{N}
\sum_{n=1}^{N}
\bar{c}_{i,j}^{(n)}.
\label{eq:app_checkpoint_sr}
\end{equation}
Hence, a later checkpoint counts only if all preceding stages are completed.

We summarize real-world performance using two metrics. \textbf{Task Success Rate (TSR)} measures the fraction of trials that finish the entire task,
\begin{equation}
\small
\mathrm{TSR}_{i}
=
\mathrm{SR}_{i,S_i},
\label{eq:app_tsr}
\end{equation}
whereas \textbf{Cumulative Success Rate (CSR)} measures average checkpoint progress,
\begin{equation}
\small
\mathrm{CSR}_{i}
=
\frac{1}{S_i}
\sum_{j=1}^{S_i}
\mathrm{SR}_{i,j}.
\label{eq:app_csr}
\end{equation}
TSR therefore reflects complete task execution, while CSR additionally captures how far unsuccessful trials progress before failure. Aggregate suite- and platform-level results are computed by averaging the corresponding task-level metrics.

\subsection{Task Descriptions}
\label{app:real_tasks}

We deploy \method{} on GALAXEA R1 Lite and AgileX Cobot Magic and evaluate the same six sequential manipulation tasks on both platforms. The suite spans repeated object sorting, container interaction, articulated-object manipulation, retrieval, and contact-rich wiping, jointly testing visual grounding, precise grasping and placement, phase transitions, and robustness to accumulated execution error. Each task is evaluated through the ordered progressive checkpoints below, with identical semantic completion criteria across the two embodiments throughout evaluation.

\begin{itemize}

\item \textbf{Block Sorting.}
\textit{Instruction: ``Place the red and blue blocks into the target bowl.''}
This task evaluates repeated object-centric pick-and-place over a multi-stage sequence, requiring the policy to preserve color identity, maintain the prescribed manipulation order, and recover a reliable grasp after completing the first placement. The four progressive stages are: 1) \textit{Pick Red Block}---securely grasping and lifting the red block; 2) \textit{+Place Red in Bowl}---placing the red block inside the target bowl; 3) \textit{+Pick Blue Block}---locating and grasping the blue block after the scene has changed; and 4) \textit{+Place Blue in Bowl}---placing the blue block inside the target bowl to complete the task.

\item \textbf{Pot Lid Retrieval.}
\textit{Instruction: ``Remove the lid and retrieve the object from the pot onto the plate.''}
This task tests interaction with a covered container and the transition from opening to retrieval. It requires the policy to remove the lid without destabilizing the pot, visually re-ground the previously occluded target, and transfer it to a constrained destination. The two progressive stages are: 1) \textit{Remove Lid}---grasping and moving the lid until the pot is accessible; and 2) \textit{+Retrieve Object to Plate}---extracting the target object and placing it stably within the designated area of the plate.

\item \textbf{Drawer Manipulation.}
\textit{Instruction: ``Open the drawer, place the toy inside, and close the drawer.''}
This task combines articulated-object control with object transport and evaluates whether a policy can maintain task coherence across distinct interaction modes. The four progressive stages are: 1) \textit{Open Drawer}---grasping the handle and opening the drawer sufficiently; 2) \textit{+Grasp Toy}---releasing the handle and securely grasping the toy; 3) \textit{+Place Toy Inside}---placing the toy fully inside the open drawer; and 4) \textit{+Close Drawer}---closing the drawer to its terminal state without displacing the inserted toy.

\item \textbf{Drawer Toy Retrieval.}
\textit{Instruction: ``Open the drawer, retrieve the toy, place it on the plate, and close the drawer.''}
This complementary drawer task tests target discovery under partial occlusion, collision-aware extraction from a confined space, and restoration of the articulated object after retrieval. The four progressive stages are: 1) \textit{Open Drawer}---opening the drawer until the interior is accessible; 2) \textit{+Retrieve Toy}---grasping and extracting the toy from the drawer; 3) \textit{+Place Toy on Plate}---transferring the toy onto the plate; and 4) \textit{+Close Drawer}---returning the drawer to its closed state.

\item \textbf{Table Wiping.}
\textit{Instruction: ``Grasp the cloth and wipe the table.''}
This contact-rich task evaluates deformable-object grasping and sustained surface interaction. Beyond reaching the cloth, the policy must establish a stable grasp, bring the cloth into contact with the tabletop, and execute a coherent wiping trajectory over the designated region. The two progressive stages are: 1) \textit{Grasp Cloth}---securely picking up the cloth; and 2) \textit{+Wipe Table}---using the grasped cloth to complete the required wiping motion while maintaining effective surface contact throughout the designated trajectory.

\item \textbf{Color Sorting.}
\textit{Instruction: ``Sort the red, blue, and yellow blocks into the matching bins.''}
This task evaluates language-conditioned color grounding and repeated placement over a long action sequence. The policy must preserve the association between each object and its color-matched receptacle while operating in a scene that changes after every placement. The three progressive stages are: 1) \textit{Red Block $\rightarrow$ Bin}---placing the red block into the matching bin; 2) \textit{+Blue Block $\rightarrow$ Bin}---subsequently placing the blue block into its matching bin; and 3) \textit{+Yellow Block $\rightarrow$ Bin}---placing the yellow block into its matching bin, thereby completing the full three-object sorting sequence.

\end{itemize}

\subsection{Qualitative Real-World Rollouts}
\label{app:real_rollouts}

Fig.~\ref{fig:real-rollouts} provides a compact view of representative
successful rollouts on GALAXEA R1 Lite and AgileX Cobot Magic. The six panels
correspond to Block Sorting, Pot Lid Retrieval, Drawer Manipulation, Drawer Toy
Retrieval, Table Wiping, and Color Sorting, with the top and bottom rows showing
the two platforms and snapshots ordered from left to right. This selection
covers repeated sorting, container interaction, articulated manipulation,
object retrieval, and contact-rich wiping. In particular, the pot and drawer
sequences make re-grounding after the scene changes visible, while Table Wiping
shows sustained surface contact and the sorting tasks show repeated object
localization and placement. Together, these sequences provide a qualitative
view of how \method{} maintains coherent progress through distinct interaction
phases.

\subsection{Detailed Real-World Results and Ablations}
\label{app:real_detailed}

\begin{figure}[t]
    \vspace{-4mm}
    \includegraphics[width=1.0\textwidth]{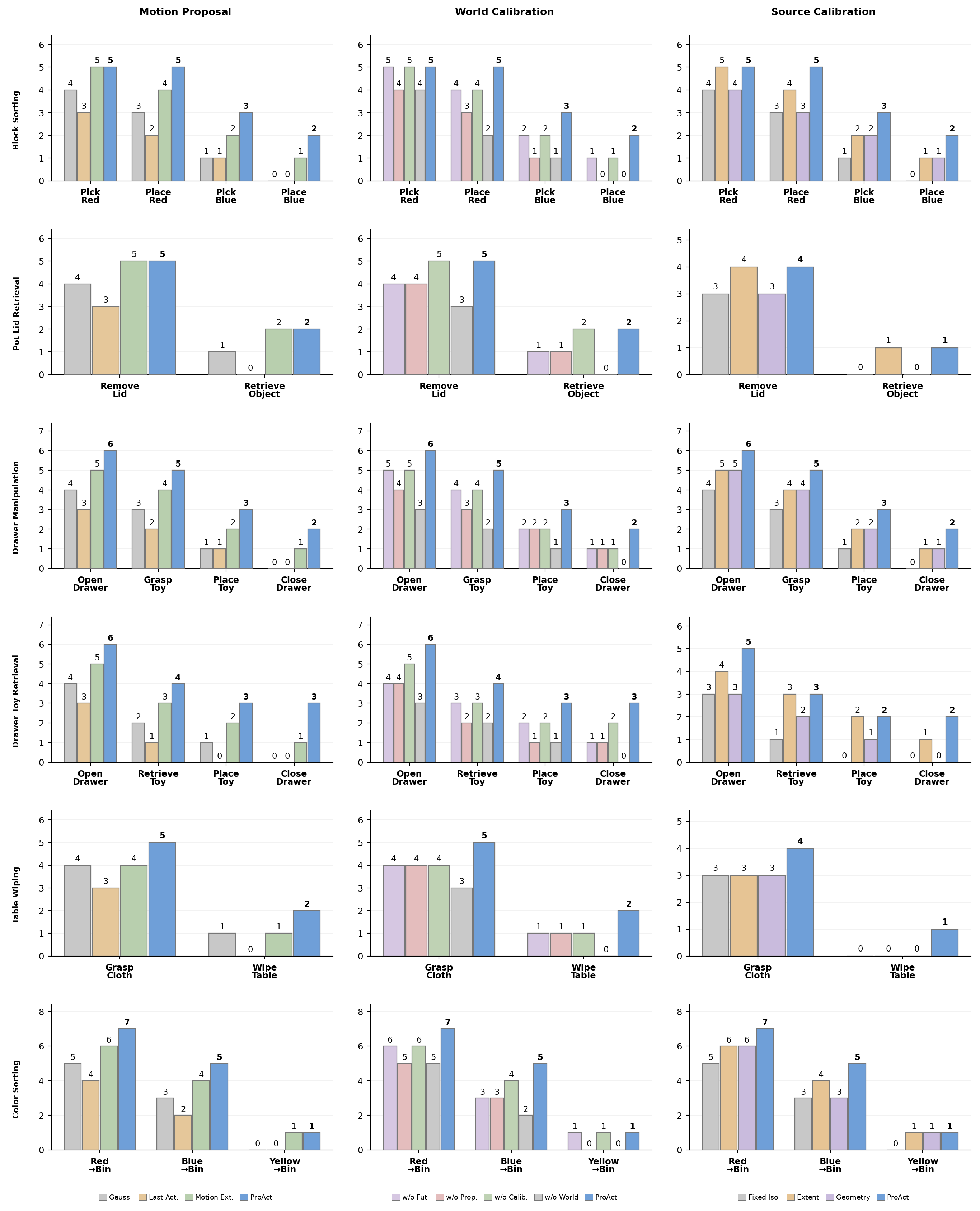}
    \vspace{-6mm}
    \caption{
    \textbf{Per-checkpoint real-world ablations across the six GALAXEA R1 Lite tasks.}
    Each panel plots per-checkpoint success for one ablation
    axis: motion proposal (left), world calibration (middle), and source
    calibration (right). Within every panel the full \method{} (blue) attains
    the highest bar, and its margin widens at the later transition and terminal
    checkpoints where local motion continuity becomes unreliable
    \textbf{across the full execution horizon}.
    }
    \label{fig:real-ablation}
    \vspace{-2mm}
\end{figure}

We complement the aggregate comparison in Section~\ref{sec:exp_real_world} with
per-checkpoint results on AgileX Cobot Magic (Tab.~\ref{tab:aloha_detailed})
and a checkpoint-level ablation on GALAXEA R1 Lite
(Fig.~\ref{fig:real-ablation}). The
progressive view isolates \emph{where} along a task the improvements and the
individual component contributions arise, which the task-level TSR/CSR
summaries cannot expose.

\begin{table}[t]
\vspace{-4mm}

\centering


\caption{
\textbf{Per-checkpoint real-world results on AgileX Cobot Magic.}
For each of the six tasks we report the number of successful trials (out of
25) at every progressive checkpoint, so that a checkpoint is credited only
when all preceding stages succeed. CSR is the mean success rate across a
task's checkpoints, and its final checkpoint equals the task TSR. Best entries
per column are in \textbf{bold}.
}
\label{tab:aloha_detailed}

\setlength{\tabcolsep}{3.0pt}
\renewcommand{\arraystretch}{1.12}


\resizebox{\textwidth}{!}{%
\begin{tabular}{l|ccccc|ccc|ccccc}
\toprule

\multirow[c]{2}{*}{%
    \raisebox{-0.7ex}{%
        \parbox[c]{1.65cm}{%
            \raggedright\textbf{Method}%
        }%
    }%
}
&
\multicolumn{5}{c|}{%
    \rule{0pt}{0.68cm}\textbf{Block Sorting}%
}
&
\multicolumn{3}{c|}{%
    \textbf{Pot Lid Retrieval}%
}
&
\multicolumn{5}{c}{%
    \textbf{Drawer Manipulation}%
}
\\[-1pt]

\cmidrule(lr){2-6}
\cmidrule(lr){7-9}
\cmidrule(lr){10-14}

&
\shortstack{Pick\\Red Block}
&
\shortstack{+Place Red\\in Bowl}
&
\shortstack{+Pick\\Blue Block}
&
\shortstack{+Place Blue\\in Bowl}
&
\textbf{CSR}

&
\shortstack{Remove\\Lid}
&
\shortstack{+Retrieve\\Object}
&
\textbf{CSR}

&
\shortstack{Open\\Drawer}
&
\shortstack{+Grasp\\Toy}
&
\shortstack{+Place Toy\\Inside}
&
\shortstack{+Close\\Drawer}
&
\textbf{CSR}
\\[-1pt]

\midrule

OpenVLA-OFT~\cite{oft}$^{\dagger}$
& 17/25
& 15/25
& 12/25
& 11/25
& 55.0\%
& 20/25
& 15/25
& 70.0\%
& 16/25
& 15/25
& 11/25
& 10/25
& 52.0\%
\\

$\pi_0$~\cite{pi0}$^{\dagger}$
& 19/25
& 17/25
& 14/25
& 13/25
& 63.0\%
& 20/25
& 14/25
& 68.0\%
& 18/25
& 16/25
& 13/25
& 12/25
& 59.0\%
\\

$\pi_{0.5}$~\cite{pi05}$^{\dagger}$
& 21/25
& 19/25
& 17/25
& 16/25
& 73.0\%
& 21/25
& 18/25
& 78.0\%
& 20/25
& 19/25
& 16/25
& 15/25
& 70.0\%
\\

\rowcolor[HTML]{DDE7F4}
\textbf{\method{}}
& \textbf{22/25}
& \textbf{21/25}
& \textbf{20/25}
& \textbf{19/25}
& \textbf{82.0\%}
& \textbf{23/25}
& \textbf{20/25}
& \textbf{86.0\%}
& \textbf{22/25}
& \textbf{21/25}
& \textbf{19/25}
& \textbf{18/25}
& \textbf{80.0\%}
\\

\bottomrule
\end{tabular}%
}

\par\vspace{1.5mm}


\resizebox{\textwidth}{!}{%
\begin{tabular}{l|ccccc|ccc|cccc}
\toprule

\multirow[c]{2}{*}{%
    \raisebox{-0.7ex}{%
        \parbox[c]{1.65cm}{%
            \raggedright\textbf{Method}%
        }%
    }%
}
&
\multicolumn{5}{c|}{%
    \rule{0pt}{0.68cm}\textbf{Drawer Toy Retrieval}%
}
&
\multicolumn{3}{c|}{%
    \textbf{Table Wiping}%
}
&
\multicolumn{4}{c}{%
    \textbf{Color Sorting}%
}
\\[-1pt]

\cmidrule(lr){2-6}
\cmidrule(lr){7-9}
\cmidrule(lr){10-13}

&
\shortstack{Open\\Drawer}
&
\shortstack{+Retrieve\\Toy}
&
\shortstack{+Place Toy\\on Plate}
&
\shortstack{+Close\\Drawer}
&
\textbf{CSR}

&
\shortstack{Grasp\\Cloth}
&
\shortstack{+Wipe\\Table}
&
\textbf{CSR}

&
\shortstack{Red\\Block$\rightarrow$Bin}
&
\shortstack{+Blue\\Block$\rightarrow$Bin}
&
\shortstack{+Yellow\\Block$\rightarrow$Bin}
&
\textbf{CSR}
\\[-1pt]

\midrule

OpenVLA-OFT~\cite{oft}$^{\dagger}$
& 14/25
& 8/25
& 8/25
& 7/25
& 37.0\%
& 19/25
& 12/25
& 62.0\%
& 20/25
& 14/25
& 9/25
& 57.3\%
\\

$\pi_0$~\cite{pi0}$^{\dagger}$
& 18/25
& 13/25
& 12/25
& 11/25
& 54.0\%
& 20/25
& 14/25
& 68.0\%
& 19/25
& 15/25
& 12/25
& 61.3\%
\\

$\pi_{0.5}$~\cite{pi05}$^{\dagger}$
& 20/25
& 16/25
& 15/25
& 14/25
& 65.0\%
& 23/25
& 18/25
& 82.0\%
& 22/25
& 19/25
& 16/25
& 76.0\%
\\

\rowcolor[HTML]{DDE7F4}
\textbf{\method{}}
& \textbf{21/25}
& \textbf{19/25}
& \textbf{18/25}
& \textbf{17/25}
& \textbf{75.0\%}
& \textbf{24/25}
& \textbf{20/25}
& \textbf{88.0\%}
& \textbf{23/25}
& \textbf{21/25}
& \textbf{18/25}
& \textbf{82.7\%}
\\

\bottomrule
\end{tabular}%
}

\end{table}
\begin{figure}[t]
    \vspace{-4mm}
    \includegraphics[width=1.0\textwidth]{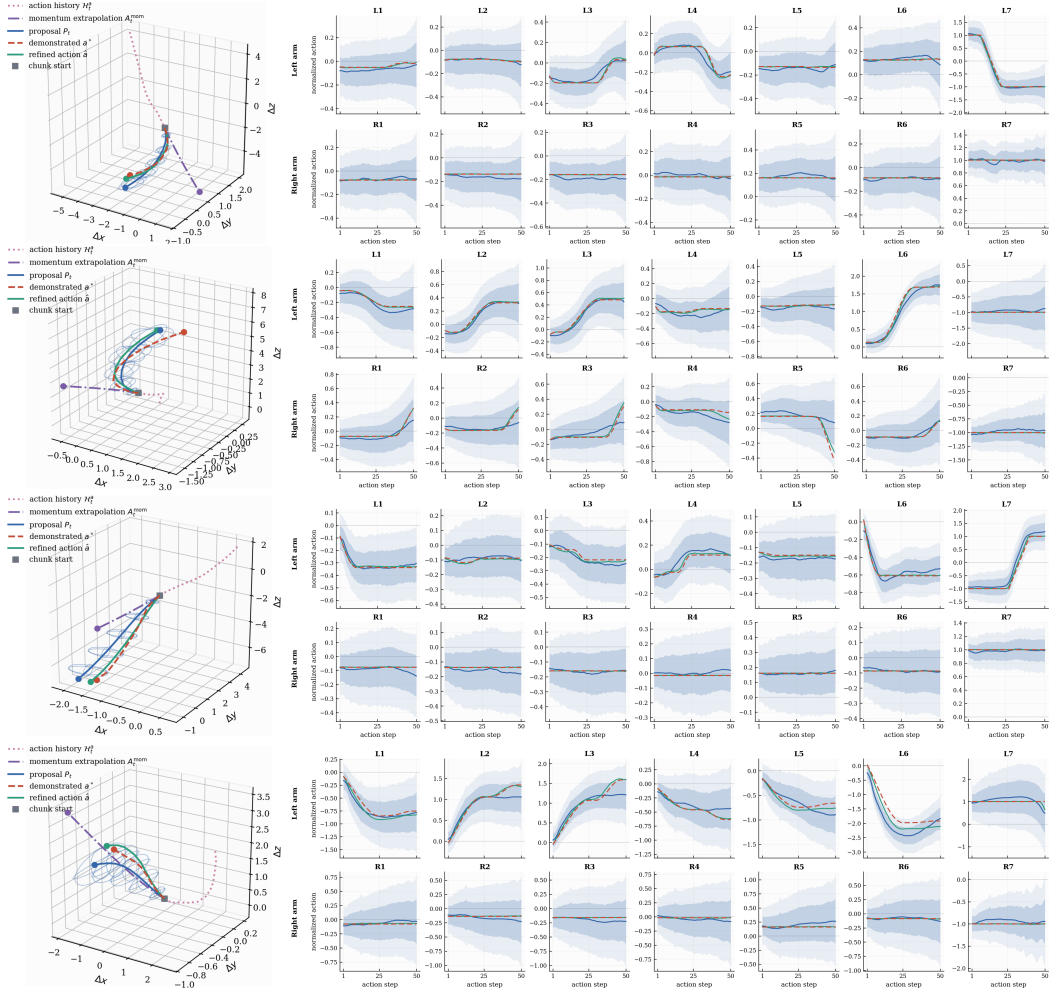}
    \vspace{-6mm}
    \caption{
    \textbf{Additional world-calibrated proposal refinements.}
    Further examples in the same format as
    Fig.~\ref{fig:vis-world}, illustrating refinement behavior across
    more cases and execution phases.
    }
    \label{fig:vis-appendix}
    \vspace{-2mm}
\end{figure}

\textbf{Gains concentrate at transition checkpoints.}
The initial acquisition checkpoints are already reliable for every method:
all policies open the drawer, remove the lid, or pick the first block on most
trials, so the advantage of \method{} accrues almost entirely at the later
stages that require re-grounding after the scene changes. On Drawer Toy
Retrieval, \method{} loses only two trials at the Open\,$\rightarrow$\,Retrieve
transition ($21\!\rightarrow\!19$), whereas $\pi_{0.5}$ loses four
($20\!\rightarrow\!16$) and OpenVLA-OFT collapses ($14\!\rightarrow\!8$); the
same pattern produces the $+10$-point CSR gaps on both drawer tasks, the
largest in the suite. Because a checkpoint is credited only after all earlier
ones succeed, these mid-sequence savings compound, so \method{} carries more
trials into the terminal checkpoint on all six tasks --- the effect that
narrows the CSR--TSR gap reported in the main text.

\textbf{Fewer late-stage failures.}
Read as survival curves, the \method{} rows decay most slowly: across the
four-checkpoint tasks it forfeits at most three trials from the first to the
last checkpoint (e.g.\ Block Sorting $22\!\rightarrow\!19$), against five for
$\pi_{0.5}$ ($21\!\rightarrow\!16$) and six or more for the weaker baselines.
The terminal contact-rich stages benefit most --- +Close Drawer, +Wipe Table,
and +Yellow$\rightarrow$Bin each improve while their preceding stages are
near-saturated --- consistent with calibration acting precisely when the
proposal must depart from recent motion.

\textbf{Components matter at the transitions, not the grasps.}
Fig.~\ref{fig:real-ablation} decomposes each ablation axis by checkpoint.
Removing any single component leaves the first checkpoint of every task
essentially unchanged --- initial grasps and openings survive --- but erodes
the later checkpoints, so the aggregate TSR drop is almost entirely a
terminal-checkpoint effect \textbf{over the full task horizon}. The World Expert is the exception: its removal
already depresses the earliest checkpoints and then decays fastest, which is
why it is the only ablation to lower performance across the whole sequence
\textbf{from early to late stages} and
yields the largest aggregate loss ($-8.0$ points in
Tab.~\ref{tab:world_ablation}).

\textbf{Extent carries more of the real-world gain.}
Within the source-calibration axis, both partial variants match \method{} on
the initial grasps but trail it at the later checkpoints, and Extent Only ends
higher than Geometry Only ($76.0$ vs.\ $74.7$ real-world SR in
Tab.~\ref{tab:source_ablation}). Adaptive magnitude
therefore accounts for more of the real-robot improvement than low-rank
geometry \textbf{throughout the later execution stages}, while only their combination recovers \method{} across the full
checkpoint sequence --- the checkpoint-level counterpart of the complementary
roles isolated in that table.

\section{Extended Qualitative Analysis}
\label{app:quantitative_analysis}

\textbf{Calibration adapts to each case.}
Fig.~\ref{fig:vis-appendix} collects several additional refinements in the
format of Fig.~\ref{fig:vis-world}, drawn from different execution phases
\textbf{and diverse interaction conditions}. The
same pattern recurs across all of them: the calibrated support is anisotropic
and phase-dependent, widening on the steps and dimensions where the proposal
(blue) must leave the momentum prior and separate from the demonstration
(dashed red), and tightening once the two agree.

\textbf{Refinement moves only where it must.}
The refined trajectory (green) stays on the proposal where recent motion is
already reliable and bends toward the demonstration only inside the widened
support. This consistency across more examples and execution phases
\textbf{under varying manipulation conditions} corroborates
the quantitative trends of Section~\ref{sec:ablation}, indicating that the
calibration responds to the local reliability of recent motion rather than
applying a fixed correction.

\section{Derivations}
\label{app:derivations}

\paragraph{Conventions and regularity conditions.}
Normalized and raw actions satisfy
$\bar a=D_a^{-1}(a^{\rm raw}-\mu_a)$, where
$D_a=\operatorname{diag}(d_1,\ldots,d_{d_a})\succ0$; $d_a$ is the number of
effective action dimensions and $d_i$ the scale of coordinate $i$. We assume
$0\leq\beta,\lambda<1$, $\tau_u>0$,
$0<\sigma_{\min}\leq\sigma_{\max}<\infty$, $d_a>1$,
$1\leq r<d_a$, and $0\leq S_{\max}<\infty$; $H$ is even when temporal
tubelets are used. Gaussian noises are mutually independent across sources and
horizon steps unless stated otherwise. The stop-gradient $\sg$ is the identity
forward and has zero derivative backward. \Eqref{eq:proact_factorization}
suppresses the perception--language context; deterministic modules correspond
to Dirac conditional kernels.

\subsection{Three-Layer KV Fusion: Risk-Optimal Window and Weights}
\label{app:kv_fusion}

The locality and weights of
$(\widetilde K_{t,j}^{v},\widetilde V_{t,j}^{v})$ in
\Eqref{eq:proposal_joint_attention} follows from one risk. Let
\begin{equation}
Y_{t,l}
=
\operatorname{RMSNorm}\!\left([K_{t,l}^{v};V_{t,l}^{v}]\right),
\label{eq:app_joint_cache}
\end{equation}
and $c_j=\rho_p(j)$. On every admissible window we assume
\textbf{(A1)} the conditional quadratic-mean model
\begin{equation}
Y_{t,c_j+r}
=
C_{t,j}+rG_{t,j}+\frac{r^2}{2}H_{t,j}+E_{t,c_j+r},
\label{eq:app_depth_expansion}
\end{equation}
where $C,G,H$ are the local value and depth variations (a discrete working
model, not a Taylor identity); \textbf{(A2)} at fixed depth and token layout,
$\mathbb E[E_{t,l}]=0$, $\mathbb E\|E_{t,l}\|_F^2=\sigma_{t,l}^2\in(0,\infty)$,
$\mathbb E\langle E_{t,l},E_{t,r}\rangle_F=0$ for $l\ne r$, and
$\langle G_{t,j},H_{t,j}\rangle_F=0$; and \textbf{(A3)} the estimand is the
smooth context $C_{t,j}$, not the observed $Y_{t,c_j}$. An omitted deterministic
higher-order term contributes bias rather than noise. Below write
$\gamma_{t,j}=\|G_{t,j}\|_F$ and $\kappa_{t,j}=\|H_{t,j}\|_F$, suppressing
their indices when unambiguous.

For $\mathcal S_j^{(m)}=\{c_j-m,\ldots,c_j+m\}$ and
$\widehat C^{(m)}(w)=\sum_rw_rY_{t,c_j+r}$ with $\sum_rw_r=1$, (A1)--(A2)
give
\begin{equation}
\mathcal{R}(m,w)
=\mathbb{E}\|\widehat C^{(m)}(w)-C_{t,j}\|_F^2
=
\Big(\sum_{r}w_rr\Big)^{2}\gamma^2
+
\frac{1}{4}\Big(\sum_{r}w_rr^2\Big)^{2}\kappa^2
+
\sum_{r}w_r^2\sigma_{t,c_j+r}^2 ,
\label{eq:app_risk_functional}
\end{equation}
from which we derive the window at uniform $w$ and the weights at fixed $m=1$.

\textbf{Window size.}
For homoscedastic noise and $w_r=(2m+1)^{-1}$,
$\sum_rw_rr=0$ and
$\sum_{r=-m}^mr^2=m(m+1)(2m+1)/3$, hence
\begin{equation}
\mathcal{R}(m)
=
\frac{\sigma^2}{2m+1}
+
\frac{\kappa^2m^2(m+1)^2}{36}.
\label{eq:app_window_risk}
\end{equation}
In particular,
$\mathcal R(0)=\sigma^2$,
$\mathcal R(1)=\sigma^2/3+\kappa^2/9$, and
$\mathcal R(2)=\sigma^2/5+\kappa^2$. Moreover,
\begin{equation}
\Delta(m):=\mathcal{R}(m+1)-\mathcal{R}(m)
=
-\frac{2\sigma^2}{(2m+1)(2m+3)}
+
\frac{\kappa^2(m+1)^3}{9},
\label{eq:app_risk_difference}
\end{equation}
is strictly increasing in $m$. Thus
$\mathcal R(1)<\mathcal R(0)$ iff $\kappa^2<6\sigma^2$, while
$\mathcal R(1)<\mathcal R(2)$ iff $\Delta(1)>0$ iff
$\kappa^2>3\sigma^2/20$; the latter also implies
$\Delta(m)>0$ for all $m\ge1$. Therefore, under (A1)--(A3),
\begin{equation}
\frac{3}{20}\sigma^2<\kappa^2<6\sigma^2
\quad\Longrightarrow\quad
m^{\star}=1,\;|\mathcal{S}_{j}^{\star}|=3 .
\label{eq:app_three_layer_optimal}
\end{equation}
and $m^\star=1$ is unique throughout the open interval; an endpoint gives a
tie. This is a conditional characterization, and one forward pass does not
identify $\kappa^2/\sigma^2$.

\textbf{Weights.}
At $m=1$, linear exactness gives $w_{\pm1}=u$, $w_0=1-2u$ and
\begin{equation}
    u^\star=\frac{2\sigma_{t,c_j}^2}
{\kappa^2+\sigma_{t,c_j-1}^2+\sigma_{t,c_j+1}^2
+4\sigma_{t,c_j}^2}.
\end{equation}

This optimum requires unknown $\kappa$ and enforces symmetric neighbors. For
the implemented weights we therefore use, only here, \textbf{(A4)} local
stationarity: $G_{t,j}=H_{t,j}=0$ on
$\mathcal S_j=\{c_j-1,c_j,c_j+1\}$. Minimizing
$\sum_lw_l^2\sigma_{t,l}^2$ subject to $\sum_lw_l=1$ then gives
$w_l^\star\propto\sigma_{t,l}^{-2}$; the nonstationary bias is quantified next.

\textbf{Estimating the variances.}
Under (A4), let
$\bar Y_{t,j}=\frac13\sum_{l\in\mathcal S_j}Y_{t,l}$ and
$d_{t,j,l}=\|Y_{t,l}-\bar Y_{t,j}\|_F^2$. Then
\begin{equation}
\mathbb{E}[d_{t,j,l}]
=
\frac{1}{3}\sigma_{t,l}^2
+
\frac{1}{9}\sum_{r\in\mathcal{S}_{j}}\sigma_{t,r}^2 ,
\qquad\text{i.e.}\qquad
\mathbb{E}[\boldsymbol d_{t,j}]=A\,\boldsymbol\sigma_{t,j}^2,
\quad
A=\tfrac{1}{3}I_3+\tfrac{1}{9}\mathbf{1}\mathbf{1}^{\top}.
\label{eq:app_dispersion_expectation}
\end{equation}
The vectors use the order $(c_j-1,c_j,c_j+1)$. Thus raw dispersion contains a
common offset, whereas
$A^{-1}=3I_3-\frac12\mathbf1\mathbf1^\top$ yields the unbiased
method-of-moments estimator
\begin{equation}
\widehat\sigma_{t,j,l}^{2}
=
3d_{t,j,l}-\frac{1}{2}\sum_{r\in\mathcal{S}_{j}}d_{t,j,r} ,
\label{eq:app_variance_moment_estimator}
\end{equation}
under (A2),(A4). Under the quadratic mean (A1), its exact bias is
\begin{equation}
    \mathbb{E}\!\left[
\widehat{\boldsymbol\sigma}_{t,j}^{2}
-\boldsymbol\sigma_{t,j}^{2}
\right]
=
\left(2\gamma_{t,j}^{2},
\frac14\kappa_{t,j}^{2}-\gamma_{t,j}^{2},
2\gamma_{t,j}^{2}\right)^{\top}.
\end{equation}

Hence the estimate is approximately unbiased only when these terms are small
relative to the layer variances. The resulting parameter-free fusion is
\begin{equation}
w_{t,j,l}
=
\frac{\max(\widehat\sigma_{t,j,l}^{2},\varepsilon_{t,j})^{-1}}
{\sum_{r\in\mathcal{S}_{j}}\max(\widehat\sigma_{t,j,r}^{2},\varepsilon_{t,j})^{-1}},
\qquad
[\widetilde K_{t,j}^{v};\widetilde V_{t,j}^{v}]
=
\sum_{l\in\mathcal{S}_{j}}w_{t,j,l}Y_{t,l}.
\label{eq:app_parameter_free_fusion}
\end{equation}
Here $\varepsilon_{t,j}>0$ combines a fixed fraction of mean dispersion with a
positive machine floor. Indeed, $\widehat\sigma_l^2\le0$ iff
$d_l\le\frac15\sum_{r\ne l}d_r$, while
$\sum_l\widehat\sigma_l^2=\frac32\sum_ld_l\ge0$; the floor therefore keeps all
weights finite even when all caches coincide.

For the instantiation of $(\rho_p, \mathcal{S}_j)$ actually used with $L_v=18$ and $L_p=6$, see
Appendix~\ref{app:kv_fusion_impl}.

\subsection{Calibrated Source: Well-Posedness}
\label{app:calibration}

\textbf{Degeneracy and projection.}
Suppress $(t,k)$ and write $d=d_a$. \Eqref{eq:geometry_loss} scores
$\widetilde C$ against one residual $\widehat r$. For
$u=\widehat r/\|\widehat r\|$ when $\widehat r\ne0$ (any unit $u$ otherwise),
set $B=sue_1^\top$. Then
\begin{align}
\widehat r^\top\widetilde C^{-1}\widehat r
&=\frac{\|\widehat r\|^2(d+s^2)}{d(1+s^2)}
\longrightarrow\frac{\|\widehat r\|^2}{d},\\
\log\det\widetilde C
&=d\log\frac{d}{d+s^2}+\log(1+s^2)
=O(1)-2(d-1)\log s.
\label{eq:app_geometry_degeneracy}
\end{align}
Thus the unconstrained pointwise objective is unbounded below; trace
normalization fixes scale but not conditioning. An amortized empirical head
would additionally need enough capacity to align with each residual, but the
following projection removes the degeneracy without that capacity assumption:

\begin{equation}
    \Pi(B)=B\min\!\left\{1,\frac{\sqrt{S_{\max}}}{\|B\|_F}\right\},
\qquad \Pi(0):=0,
\end{equation}

followed by $B\leftarrow\Pi(B)$, so $\|B\|_F^2\le S_{\max}$. If $s_i$ are the
singular values of $B$, then $r<d$ leaves $d-r$ unit eigenvalues and
\begin{equation}
\kappa(\widetilde C_{t,k})
=
1+\max_i s_i^2
\le
1+\sum_i s_i^2
\le
1+S_{\max}.
\label{eq:app_geometry_condition}
\end{equation}
Recomputing $\alpha_{t,k}=d/(d+\|B\|_F^2)$ after projection preserves
$\operatorname{tr}C_{t,k}=d\sigma_{t,k}^2$. Moreover,
$C_{t,k}^a=D_aC_{t,k}D_a^\top$ is positive definite and
$\kappa(C_{t,k}^a)\le\kappa(D_a)^2(1+S_{\max})$; the sharper
$1+S_{\max}$ bound applies in normalized coordinates.

\textbf{Lower bound.}
Let $S=\|B\|_F^2=\sum_i s_i^2$. Since
$\prod_i(1+s_i^2)\ge1+S$,
\begin{equation}
\begin{aligned}
\log\det\widetilde C
&= d\log\frac{d}{d+S}+\sum_i\log(1+s_i^2) \\
&\ge g(S):=d\log\frac{d}{d+S}+\log(1+S),
\qquad
g'(S)=-\frac{(d-1)S}{(d+S)(1+S)}\le 0.
\end{aligned}
\end{equation}

Using $S\le S_{\max}$ and the nonnegative Mahalanobis term gives
\begin{equation}
\mathcal{L}_B
\ge
\frac{1}{2}
\left[
d_a\log\frac{d_a}{d_a+S_{\max}}
+
\log(1+S_{\max})
\right].
\label{eq:app_geometry_lower_bound}
\end{equation}
For $d_a=7$, $S_{\max}=5$, the bound is $-0.991$. Equality requires zero
residual, $S=S_{\max}$, and one nonzero singular direction; a nonzero residual
has a strictly higher sample-wise minimum. Thus negative losses are possible,
but divergence to $-\infty$ is excluded.

For the $O(d_ar^2)$ evaluation and sampling that this projection enables, see
Appendix~\ref{app:calibration_impl}.

\section{Limitations and Future Work} 
\label{app:limitations_future}

While \method{} improves generative action prediction by calibrating motion-informed proposals with prospective world representations, several limitations remain.

\textbf{Counterfactual supervision.}
The current formulation is trained primarily from factual expert trajectories, where each action sequence is paired with the future evolution that actually occurred. As a result, the model does not explicitly observe the counterfactual world evolution that would be induced by alternative or erroneous proposals. The calibration module therefore learns how a proposal should be adjusted under expert future supervision, rather than directly comparing the predicted consequence of a proposal against the desired task-consistent future. A promising direction is to incorporate proposal perturbations, simulator rollouts, or action-conditioned world models to provide explicit counterfactual supervision and enable consequence-aware calibration.

\textbf{Fixed-horizon prospective reasoning.}
Although \method{} is supervised with a temporal future sequence rather than a single endpoint, its prospective reasoning is currently performed over a fixed horizon aligned with the action chunk. However, different manipulation stages may require different temporal scales: fine-grained contact and alignment can often be resolved from short-term evolution, whereas the consequences of decisions in long-horizon tasks may only become evident after several action chunks. Extending the current formulation toward multi-scale or adaptive-horizon world reasoning could allow the policy to calibrate proposals using both immediate interaction cues and longer-term task progress \textbf{across diverse manipulation phases and tasks}.

\end{document}